\documentclass[10pt,twocolumn,letterpaper]{article}

\usepackage{cvpr}
\usepackage[utf8]{inputenc} 
\usepackage[T1]{fontenc}    
\usepackage{times}
\usepackage{epsfig}
\usepackage{graphicx}
\usepackage{amsmath}
\usepackage{amssymb}
\usepackage{caption}
\usepackage{url}            
\usepackage{booktabs} 
\usepackage{amsfonts} 
\usepackage{nicefrac} 
\usepackage{microtype} 
\usepackage{subfig}
\usepackage{algorithm}
\usepackage{algpseudocode}
\usepackage{rotating}
\usepackage{multirow}
\usepackage{array}

\DeclareMathOperator*{\argmax}{arg\,max}
\newcommand{\norm}[1]{\left\lVert#1\right\rVert}

\usepackage[breaklinks=true,bookmarks=false,colorlinks]{hyperref}
\hypersetup{
    linkcolor=black,
}

\cvprfinalcopy 

\def\cvprPaperID{6129} 

\begin{document}

\def\equationautorefname{Eq.}
\def\figureautorefname{Fig.}
\def\sectionautorefname{Section}


\title{Decoupling Direction and Norm for Efficient Gradient-Based $L_2$ \\Adversarial Attacks and Defenses}

\author{
    J\'er\^ome Rony\thanks{Equal contribution.} ~\textsuperscript{1} \qquad Luiz G. Hafemann\footnotemark[1] ~\textsuperscript{1}\qquad Luiz S. Oliveira\textsuperscript{2}\qquad Ismail Ben Ayed\textsuperscript{1}\\ Robert Sabourin\textsuperscript{1}\qquad Eric Granger\textsuperscript{1}  \\
    \textsuperscript{1}Laboratoire d'imagerie, de vision et d'intelligence artificielle (LIVIA), \'ETS Montreal, Canada\\
    \textsuperscript{2}Department of Informatics, Federal University of Paran\'a, Curitiba, Brazil \\
    {\tt\small jerome.rony@gmail.com\qquad luiz.gh@mailbox.org\qquad lesoliveira@inf.ufpr.br}\\ {\tt\small\{ismail.benayed, robert.sabourin, eric.granger\}@etsmtl.ca}
    \vspace{-0.8em}
}

\maketitle

\begin{abstract}
Research on adversarial examples in computer vision tasks has shown that small, often imperceptible changes to an image can induce misclassification, which has security implications for a wide range of image processing systems. Considering $L_2$ norm distortions, the Carlini and Wagner attack is presently the most effective white-box attack in the literature. However, this method is slow since it performs a line-search for one of the optimization terms, and often requires thousands of iterations. In this paper, an efficient approach is proposed to generate gradient-based attacks that induce misclassifications with low $L_2$ norm, by decoupling the direction and the norm of the adversarial perturbation that is added to the image. Experiments conducted on the MNIST, CIFAR-10 and ImageNet datasets indicate that our attack achieves comparable results to the state-of-the-art (in terms of $L_2$ norm) with considerably fewer iterations (as few as 100 iterations), which opens the possibility of using these attacks for adversarial training. Models trained with our attack achieve state-of-the-art robustness against white-box gradient-based $L_2$ attacks on the MNIST and CIFAR-10 datasets, outperforming the Madry defense when the attacks are limited to a maximum norm.
\end{abstract}

\section{Introduction}

Deep neural networks have achieved state-of-the-art performances on a wide variety of computer vision applications, such as image classification, object detection, tracking, and activity recognition~\cite{gu_recent_2018}. In spite of their success in addressing these challenging tasks, they are vulnerable to active \emph{adversaries}. Most notably, they are susceptible to \emph{adversarial examples}\footnote{This also affects other machine learning classifiers, but we restrict our analysis to CNNs, that are most commonly used in computer vision tasks.}, in which adding small perturbations to an image, often imperceptible to a human observer, causes a misclassification \cite{biggio_wild_2017, szegedy_intriguing_2013}. 

Recent research on adversarial examples developed \emph{attacks} that allow for evaluating the robustness of models, as well as \emph{defenses} against these attacks. Attacks have been proposed to achieve different objectives, such as minimizing the amount of noise that induces misclassification \cite{carlini_towards_2017, szegedy_intriguing_2013}, or being fast enough to be incorporated into the training procedure \cite{goodfellow_explaining_2014, tramer_ensemble_2017}. In particular, considering the case of obtaining adversarial examples with lowest perturbation (measured by its $L_2$ norm), the state-of-the-art attack has been proposed by Carlini and Wagner (C\&W) \cite{carlini_towards_2017}. While this attack generates adversarial examples with low $L_2$ noise, it also requires a high number of iterations, which makes it impractical for training a robust model to defend against such attacks. In contrast, one-step attacks are fast to generate, but using them for training does not increase model robustness on white-box scenarios, with full knowledge of the model under attack \cite{tramer_ensemble_2017}. Developing an attack that finds adversarial examples with low noise in few iterations would enable adversarial training with such examples, which could potentially increase model robustness against white-box attacks.


Developing attacks that minimize the norm of the adversarial perturbations requires optimizing two objectives: 1) obtaining a low $L_2$ norm, while 2) inducing a misclassification. 
With the current state-of-the-art method (C\&W \cite{carlini_towards_2017}), this is addressed by using a two-term loss function, with the weight balancing the two competing objectives found via an expensive line search, requiring a large number of iterations. This makes the evaluation of a system's robustness very slow and it is unpractical for adversarial training.


In this paper, we propose an efficient gradient-based attack called \emph{Decoupled Direction and Norm}\footnote{Code available at \url{https://github.com/jeromerony/fast_adversarial}.} (DDN) that induces misclassification with a low $L_2$ norm.  This attack optimizes the cross-entropy loss, and instead of penalizing the norm in each iteration, projects the perturbation onto a $L_2$-sphere centered at the original image. The change in norm is then based on whether the sample is adversarial or not. Using this approach to decouple the direction and norm of the adversarial noise leads to an attack that needs significantly fewer iterations, achieving a level of performance comparable to state-of-the-art, while being amenable to be used for adversarial training.

A comprehensive set of experiments was conducted using the MNIST, CIFAR-10 and ImageNet datasets. Our attack obtains comparable results to the state-of-the-art while requiring much fewer iterations (\textasciitilde100 times less than C\&W). For untargeted attacks on the ImageNet dataset, our attack achieves better performance than the C\&W attack, taking less than 10 minutes to attack 1\,000 images, versus over 35 hours to run the C\&W attack.

Results for adversarial training on the MNIST and CIFAR-10 datasets indicate that DDN can achieve  state-of-the-art robustness compared to the Madry defense \cite{madry_towards_2017}. These models require that attacks use a higher average $L_2$ norm to induce misclassifications. They also obtain a higher accuracy when the $L_2$ norm of the attacks is bounded. On MNIST, if the attack norm is restricted to $1.5$, the model trained with the Madry defense achieves 67.3\% accuracy, while our model achieves 87.2\% accuracy. On CIFAR-10, for attacks restricted to a norm of $0.5$, the Madry model achieves 56.1\% accuracy, compared to 67.6\% in our model.



\section{Related Work}

In this section, we formalize the problem of adversarial examples, the threat model, and review the main attack and defense methods proposed in the literature.

\subsection{Problem Formulation} 

\begin{figure}
    \centering
    \includegraphics[width=\columnwidth]{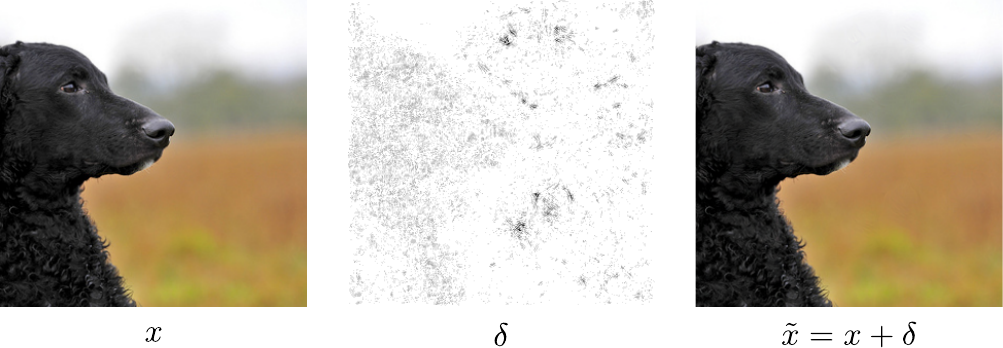}
    \caption{Example of an adversarial image on the ImageNet dataset. The sample $x$ is recognized as a Curly-coated retriever. Adding a perturbation $\delta$ we obtain an adversarial image that is classified as a microwave (with $\norm{\delta}_2 = 0.7$).}
    \label{fig:adv_example}
    \vspace{-1em}
\end{figure}

Let $x$ be an sample from the input space $\mathcal{X}$, with label $y_\text{true}$ from a set of possible labels $\mathcal{Y}$. Let $D(x_1, x_2)$ be a distance measure that compares two input samples (ideally capturing their perceptual similarity). $P(\text{y}|x, \theta)$ is a model (classifier) parameterized by $\theta$. An example $\tilde{x} \in \mathcal{X}$ is called \emph{adversarial} (for non-targeted attacks) against the classifier if $\argmax_j P(y_j | \tilde{x}, \theta) \neq y_\text{true}$ and $D(x, \tilde{x}) \le \epsilon$, for a given maximum perturbation $\epsilon$. A \emph{targeted attack} with a given desired class $y_\text{target}$ further requires that $\argmax_j P(y_j | \tilde{x}, \theta) = y_\text{target}$. We denote as $J(x, y, \theta)$, the cross-entropy between the prediction of the model for an input $x$ and a label $y$.
\autoref{fig:adv_example} illustrates a targeted attack on the ImageNet dataset, against an Inception v3 model \cite{szegedy_rethinking_2015}. 

In this paper, attacks are considered to be generated by a gradient-based optimization procedure, restricting our analysis to differentiable classifiers. These attacks can be formulated either to obtain a minimum distortion $D(x, \tilde{x})$, or to obtain the worst possible loss in a region $D(x, \tilde{x}) \le \epsilon$. As an example, consider that the distance function is a norm (e.g.,  $L_0$, $L_2$ or $L_\infty$), and the inputs are images (where each pixel's value is constrained between 0 and $M$). In a white-box scenario, the optimization procedure to obtain an non-targeted attack with minimum distortion $\delta$  can be formulated as:
\begin{equation}
\begin{aligned}
\min_\delta \norm{\delta} \;\; \text{subject to} \;\; &\argmax_j{P(y_j| x + \delta, \theta)} \neq y_\text{true} \\ 
\text{and} \;\; &\, 0 \leq x + \delta \leq M
\end{aligned}
\label{eq:closer_adv}
\end{equation}
With a similar formulation for \emph{targeted attacks}, by changing the constraint to be equal to the target class.

If the objective is to obtain the worst possible loss for a given maximum noise of norm $\epsilon$, the problem can be formulated as:
\begin{equation}
\begin{aligned}
\min_\delta P(y_\text{true} | x + \delta, \theta) \;\; \text{subject to} \;\; & \norm{\delta} \le \epsilon \\ 
 \text{and} \;\; &\, 0 \leq x + \delta \leq M
\end{aligned}
\label{eq:worst_adv}
\end{equation}
With a similar formulation for \emph{targeted attacks}, by maximizing $P(y_\text{target} | x + \delta, \theta)$.

We focus on gradient-based attacks that optimize the $L_2$ norm of the distortion. While this distance does not perfectly capture perceptual similarity, it is widely used in computer vision to measure similarity between images (e.g. comparing image compression algorithms, where Peak Signal-to-Noise Ratio is used, which is directly related to the $L_2$ measure). A differentiable distance measure that captures perceptual similarity is still an open research problem.

\subsection{Threat Model}

In this paper, a \emph{white-box} scenario is considered, also known as a Perfect Knowledge scenario \cite{biggio_wild_2017}. In this scenario, we consider that an attacker has perfect knowledge of the system, including the neural network architecture and the learned weights $\theta$. This threat model serves to evaluate system security under the \emph{worst case} scenario. Other scenarios can be conceived to evaluate attacks under different assumptions on the attacker's knowledge, for instance, no access to the trained model, no access to the same training set, among others. These scenarios are referred as \emph{black-box} or Limited-Knowledge \cite{biggio_wild_2017}.


\subsection{Attacks}

Several attacks were proposed in the literature, either focusing on obtaining adversarial examples with a small $\delta$ (\autoref{eq:closer_adv}) \cite{ carlini_towards_2017, moosavi2016deepfool, szegedy_intriguing_2013}, or on obtaining adversarial examples in one (or few) steps for adversarial training \cite{goodfellow_explaining_2014, kurakin_adversarial_2016-1}. 

\textbf{L-BFGS.} 
Szegedy \etal \cite{szegedy_intriguing_2013} proposed an attack for minimally distorted examples  (\autoref{eq:closer_adv}), by considering the following approximation:
\begin{equation}
\begin{aligned}
\min_\delta C \norm{\delta}_2 + \log P(y_\text{true}| x + \delta, \theta)  \;\; \\ \text{subject to} \;\;  0 \leq x + \delta \leq M
\end{aligned}
\label{eq:lbfgs}
\end{equation}
where the constraint $x+\delta \in [0, M]^n$ was addressed by using a box-constrained optimizer (L-BFGS: Limited memory Broyden–Fletcher–Goldfarb–Shanno), and a line-search to find an appropriate value of $C$.
 
\textbf{FGSM.} 
Goodfellow \etal \cite{goodfellow_explaining_2014} proposed the Fast Gradient Sign Method, a one-step method that could generate adversarial examples. The original formulation was developed considering the $L_\infty$ norm, but it has also been used to generate attacks that focus on the $L_2$ norm as follows:
\begin{equation}
    \begin{gathered}
        \tilde{x} = x + \epsilon \frac{\nabla_x J(x, y, \theta)}{\norm{\nabla_x J(x, y, \theta)}}
    \end{gathered}
\end{equation}
where the constraint $\tilde{x} \in [0, M]^n$ was addressed by simply clipping the resulting adversarial example. 

\textbf{DeepFool.}
This method considers a linear approximation of the model, and iteratively refines an adversary example by choosing the point that would cross the decision boundary under this approximation. This method was developed for untargeted attacks, and for any $L_p$ norm \cite{moosavi2016deepfool}.


\textbf{C\&W.} 
Similarly to the L-BFGS method, the C\&W $L_2$ attack \cite{carlini_towards_2017} minimizes two criteria at the same time -- the perturbation that makes the sample adversarial (e.g., misclassified by the model), and the $L_2$ norm of the perturbation. Instead of using a box-constrained optimization method, they propose changing variables using the $\tanh$ function, and instead of optimizing the cross-entropy of the adversarial example, they use a difference between logits. For a targeted attack aiming to obtain class $t$, with $Z$  denoting the model output before the softmax activation (logits), it optimizes:
\begin{equation}
    \begin{gathered}
        \min_\delta \Big[\| \tilde{x} - x \|_{2}^{2} + C f(\tilde{x}) \Big]\\
        \begin{aligned}
            \text{where} \quad f(\tilde{x}) &= \max (\max_{i \neq t} \{ Z(\tilde{x})_i\} - Z(\tilde{x})_{t}, -\kappa)\\
            \text{and} \quad \tilde{x} &= \frac{1}{2}(\tanh(\mathrm{arctanh}(x) + \delta) + 1)
        \end{aligned}
    \end{gathered}
\label{eq:carlini}
\end{equation}
where $Z(\tilde{x})_{i}$ denotes the logit corresponding to the $i$-th class. By increasing the confidence parameter $\kappa$, the adversarial sample will be misclassified with higher confidence. To use this attack in the untargeted setting, the definition of $f$ is modified to ${f(\tilde{x}) = \max ( Z(\tilde{x})_y - \max_{i \neq y} \{Z(\tilde{x})_i\}, -\kappa)}$ where $y$ is the original label. 

\subsection{Defenses}

Developing defenses against adversarial examples is an active area of research. To some extent, there is an \emph{arms race} on developing defenses and attacks that break them. Goodfellow \etal proposed a method called \emph{adversarial training} \cite{goodfellow_explaining_2014}, in which the training data is augmented with FGSM samples. This was later shown not to be robust against iterative white-box attacks, nor black-box single-step attacks \cite{tramer_ensemble_2017}. Papernot \etal \cite{papernot_distillation_2016} proposed a \emph{distillation} procedure to train robust networks, which was shown to be easily broken by iterative white-box attacks \cite{carlini_towards_2017}. Other defenses involve  \emph{obfuscated gradients} \cite{athalye_obfuscated_2018}, where models either incorporate non-differentiable steps (such that the gradient cannot be computed) \cite{buckman_thermometer_2018, guo_countering_2018}, or randomized elements (to induce incorrect estimations of the gradient) \cite{dhillon_stochastic_2018, xie_mitigating_2018}. These defenses were later shown to be ineffective when attacked with Backward Pass Differentiable Approximation (BPDA) \cite{athalye_obfuscated_2018}, where the actual model is used for forward propagation, and the gradient in the backward-pass is approximated. The Madry defense \cite{madry_towards_2017}, which considers a worst-case optimization, is the only defense that has been shown to be somewhat robust (on the MNIST and CIFAR-10 datasets). Below we provide more detail on the general approach of adversarial training, and the Madry defense.

\textbf{Adversarial Training.}
This defense considers augmenting the training objective with adversarial examples \cite{goodfellow_explaining_2014}, with the intention of improving robustness. Given a model with loss function $J(x, y, \theta)$, training is augmented as follows:
\begin{equation}
    \tilde{J}(x, y, \theta) = \alpha J(x, y, \theta) + (1-\alpha) J(\tilde{x}, y, \theta)
\end{equation}
where $\tilde{x}$ is an adversarial sample. In \cite{goodfellow_explaining_2014}, the FGSM is used to generate the adversarial example in a single step. Tram\`er \etal \cite{tramer_ensemble_2017} extended this method, showing that generating one-step attacks using the model under training introduced an issue. The model can converge to a degenerate solution where its gradients produce ``easy'' adversarial samples, causing the adversarial loss to have a limited influence on the training objective. They proposed a method in which an ensemble of models is also used to generate the adversarial examples $\tilde{x}$. This method displays some robustness against black-box attacks using surrogate models, but does not increase robustness in white-box scenarios.

\textbf{Madry Defense.}
Madry \etal \cite{madry_towards_2017} proposed a saddle point optimization problem, optimizing for the worst case:
\begin{equation}
\begin{gathered}
    \min_\theta p(\theta) \\
    \text{where} \quad p(\theta) = \mathbb{E}_{(x,y) \sim \mathcal{D}}\big[\max_{\delta \in \mathcal{S}} J(x + \delta, y, \theta) \big]
\end{gathered}
\label{eq:madry}
\end{equation}
where $\mathcal{D}$ is the training set, and $\mathcal{S}$ indicates the feasible region for the attacker (e.g. $\mathcal{S} = \{\delta : \norm{\delta} < \epsilon\}$). They show that \autoref{eq:madry} can be optimized by stochastic gradient descent -- during each training iteration, it first finds the adversarial example that maximizes the loss around the current training sample $x$ (i.e., maximizing the loss over $\delta$, which is equivalent to minimizing the probability of the correct class as in \autoref{eq:worst_adv}), and then, it minimizes the loss over $\theta$. Experiments in Athalye \etal. \cite{athalye_obfuscated_2018} show that it was the only defense not broken under white-box attacks.

\section{Decoupled Direction and Norm Attack}
\label{sec:attack}

\begin{figure}
    \centering
    \includegraphics[scale=0.5]{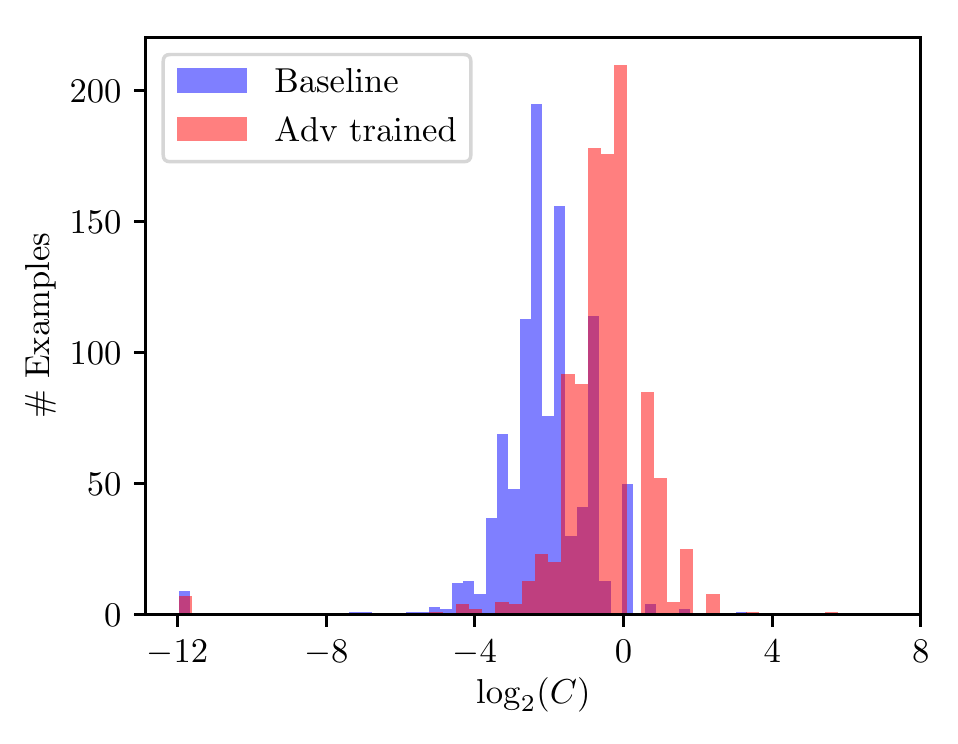}
    \caption{Histogram of the best $C$ found by the C\&W algorithm with 9 search steps on the MNIST dataset.}
    \label{fig:mnist_best_c}
    \vspace{-1em}
\end{figure}

From the problem definition, we see that finding the worst adversary in a fixed region is an easier task. In \autoref{eq:worst_adv}, both constraints can be expressed in terms of $\delta$, and the resulting equation can be optimized using projected gradient descent. Finding the closest adversarial example is harder: \autoref{eq:closer_adv} has a constraint on the prediction of the model, which cannot be addressed by a simple projection. A common approach, which is used by Szegedy \etal \cite{szegedy_intriguing_2013} and in the C\&W \cite{carlini_towards_2017} attack, is to approximate the constrained problem in \autoref{eq:closer_adv} by an unconstrained one, replacing the constraint with a {\em penalty}. This amounts to jointly optimizing both terms, the norm of $\delta$ and a classification term (see Eq. \ref{eq:lbfgs} and \ref{eq:carlini}), with a sufficiently high parameter $C$. In the general context of constrained optimization, such a penalty-based approach is a well known general principle \cite{Jensen_Bard_2003}. While tackling an unconstrained problem is convenient, penalty methods have well-known difficulties in practice. The main difficulty is that one has to choose parameter $C$ in an {\em ad hoc} way. For instance, if $C$ is too small in \autoref{eq:carlini}, the example will not be adversarial; if it is too large, this term will dominate, and result in an adversarial example with more noise. This can be particularly problematic when optimizing with a low number of steps (e.g. to enable its use in adversarial training). \autoref{fig:mnist_best_c} plots a histogram of the values of $C$ that were obtained by running the C\&W attack on the MNIST dataset. We can see that the optimum $C$ varies significantly among different examples, ranging from $2^{-11}$ to $2^{5}$. We also see that the distribution of the best constant $C$ changes whether we attack a model with or without adversarial training (adversarially trained models often require higher $C$). Furthermore, penalty methods typically result in slow convergence \cite{Jensen_Bard_2003}. 

\begin{algorithm}
    \caption{Decoupled Direction and Norm Attack}
    \label{alg:ddn}
    \begin{algorithmic}[1] 
        \Require $x$: original image to be attacked
        \Require $y$: true label (untargeted) or target label (targeted)
        \Require $K$: number of iterations
        \Require $\alpha$: step size
        \Require $\gamma$: factor to modify the norm in each iteration
        \Ensure $\tilde{x}$: adversarial image
        \State Initialize $\delta_0 \gets \textbf{0}$, $\tilde{x}_0 \gets x$, $\epsilon_0 \gets 1$
        \State If targeted attack: $m \gets -1$ else $m \gets +1$
        \For{$k \gets 1 $ to $K$}
            \State $g \gets m \nabla_{\tilde{x}_{k-1}}J(\tilde{x}_{k-1}, y, \theta)$
            \State $g \gets \alpha \frac{g}{\norm{g}_2}$ \Comment{\parbox[t]{.33\linewidth}{Step of size $\alpha$ in the direction of $g$}}
            \State $\delta_k \gets \delta_{k-1} + g$
            \If{$\tilde{x}_{k-1}$ is adversarial} 
                \State $\epsilon_k \gets (1-\gamma) 
                \epsilon_{k-1}$ \Comment{Decrease norm}
            \Else
                \State $\epsilon_k \gets (1+\gamma)
                \epsilon_{k-1}$ \Comment{Increase norm}
            \EndIf
            \State $\tilde{x}_k \gets x + \epsilon_k \frac{\delta_k}{\norm{\delta_k}_2}$ 
            \Comment{\parbox[t][2.3em]{.35\linewidth}{Project $\delta_k$ onto an\\ $\epsilon_k$-sphere around $x$}}
            \State $\tilde{x}_k \gets \text{clip}(\tilde{x}_k, 0, 1)$ \Comment{Ensure $\tilde{x}_k \in \mathcal{X}$}
        \EndFor
        \State Return $\tilde{x}_k$ that has lowest norm $\norm{\tilde{x}_k - x}_2$ and is adversarial
    \end{algorithmic}
\end{algorithm}

\begin{figure}
    \centering
    \subfloat[$\tilde{x}_k$ not adversarial]{
    \includegraphics[scale=0.32]{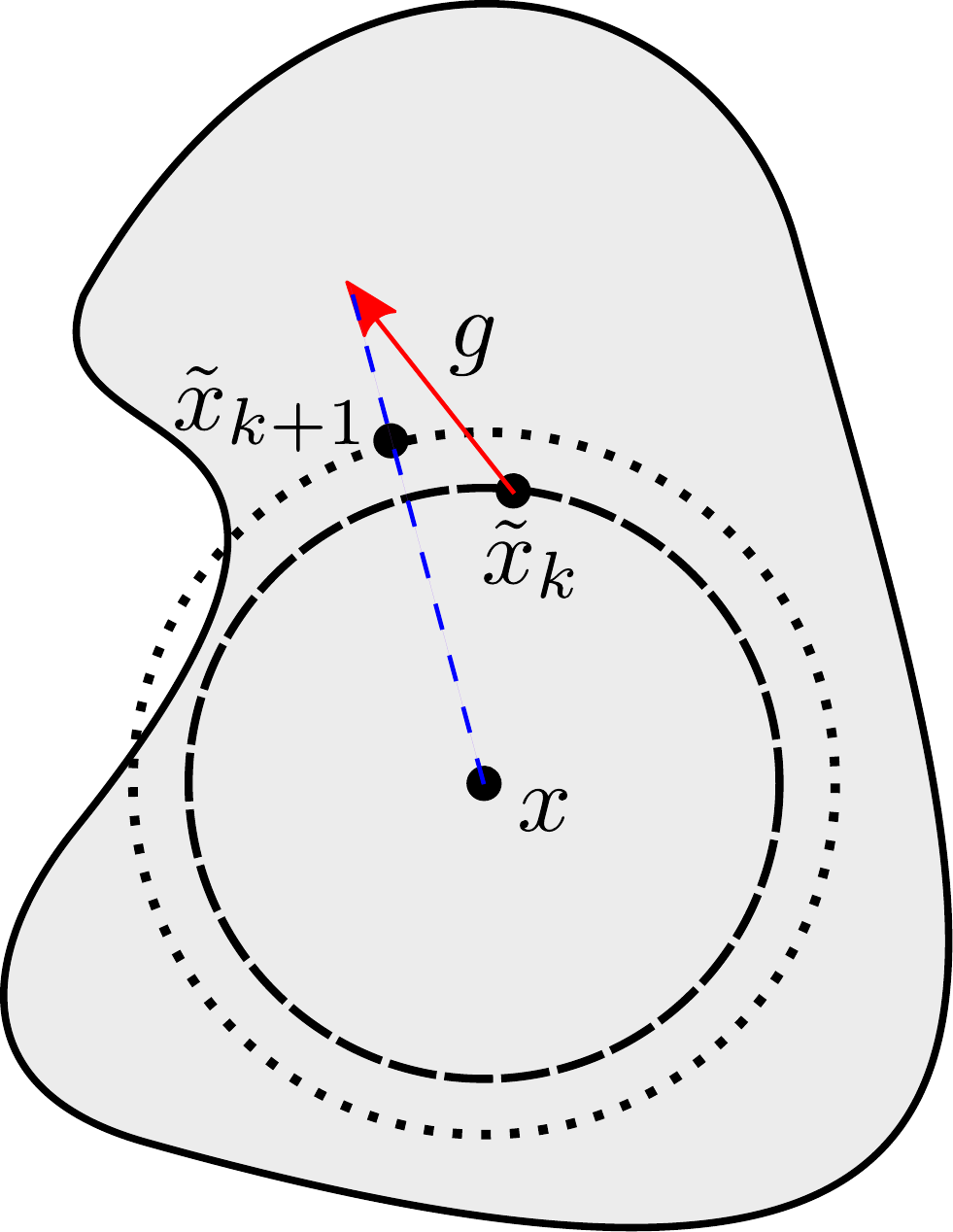}}
    \qquad
    \subfloat[$\tilde{x}_k$ adversarial]{
    \includegraphics[scale=0.32]{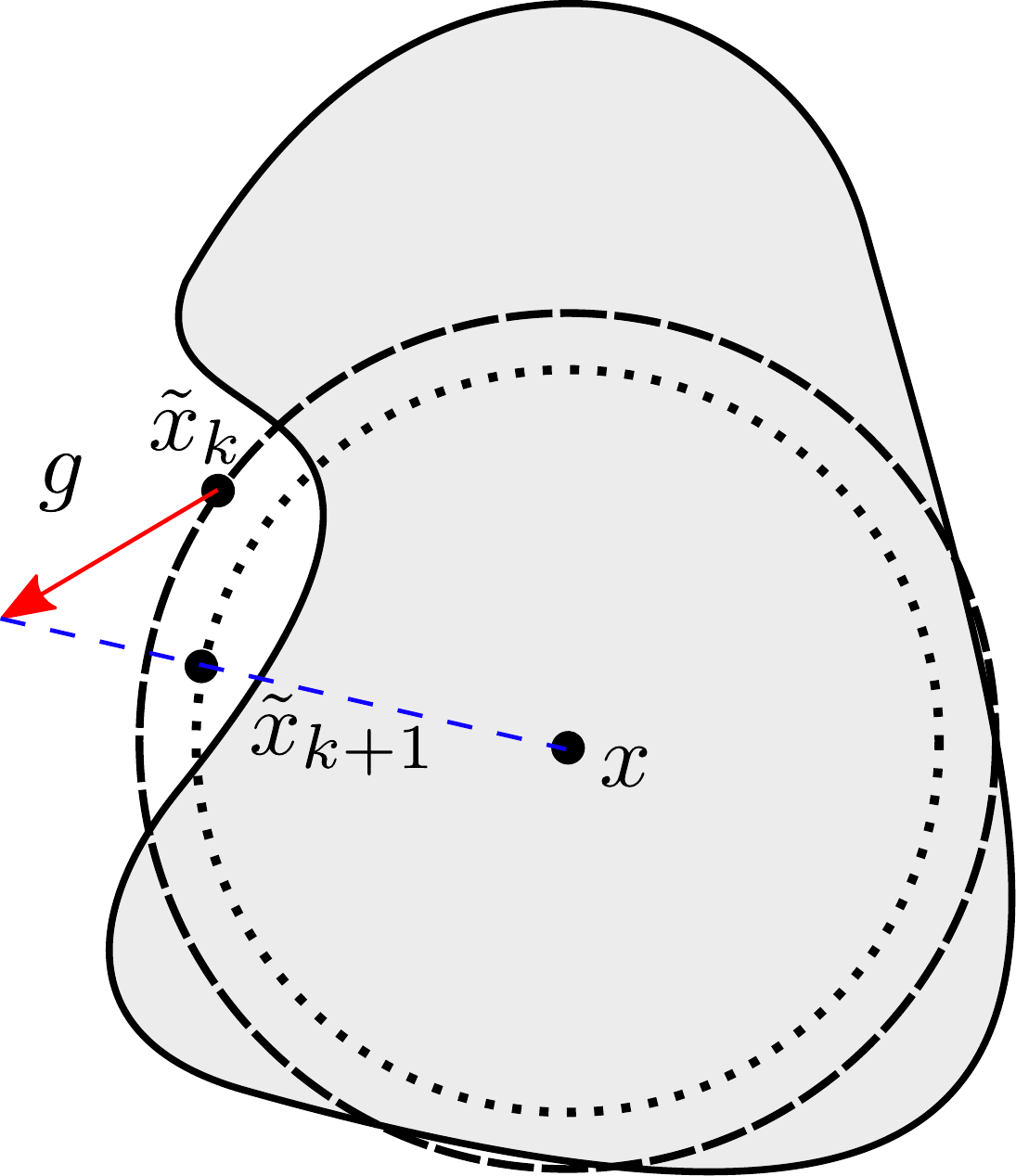}}
    \caption{Illustration of an untargeted attack. The shaded area denotes the region of the input space classified as $y_\text{true}$. In (a), $\tilde{x}_k$ is still not adversarial, and we increase the norm $\epsilon_{k+1}$ for the next iteration, otherwise it is reduced in (b). In both cases, we take a step $g$ starting from the current point $\tilde{x}$, and project back to an $\epsilon_{k+1}$-sphere centered at $x$.}
    \label{fig:ddn_illustration}
    \vspace{-1em}
\end{figure}

Given the difficulty of finding the appropriate constant $C$ for this optimization, we propose an algorithm that does not impose a penalty on the $L_2$ norm during the optimization. Instead, the norm is constrained by projecting the adversarial perturbation $\delta$ on an $\epsilon$-sphere around the original image $x$. Then, the $L_2$ norm is modified through a binary decision. If the sample $x_k$ is not adversarial at step $k$, the norm is increased for step $k+1$, otherwise it is decreased. 

We also note that optimizing the cross-entropy may present two other difficulties. First, the function is not bounded, which can make it dominate in the optimization of \autoref{eq:lbfgs}. Second, when attacking trained models, often the predicted probability of the correct class for the original image is very close to 1, which causes the cross entropy to start very low and increase by several orders of magnitude during the search for an adversarial example. This affects the norm of the gradient, making it hard to find an appropriate learning rate. C\&W address these issues by optimizing the difference between logits instead of the cross-entropy. In this work, the issue of it being unbounded does not affect the attack procedure, since the decision to update the norm is done on the model's prediction (not on the cross-entropy). In order to handle the issue of large changes in gradient norm, we normalize the gradient to have unit norm before taking a step in its direction.


The full procedure is described in \autoref{alg:ddn} and illustrated in \autoref{fig:ddn_illustration}. We start from the original image $x$, and iteratively refine the noise $\delta_k$. In iteration $k$, if the current sample $\tilde{x}_k = x + \delta_k$ is still not adversarial, we consider a larger norm $\epsilon_{k+1} = (1+\gamma)\epsilon_k$. Otherwise, if the sample is adversarial, we consider a smaller $\epsilon_{k+1} = (1-\gamma)\epsilon_k$. 
In both cases, we take a step $g$ (step 5 of \autoref{alg:ddn}) from the point $\tilde{x}_k$ (red arrow in \autoref{fig:ddn_illustration}), and project it back onto an $\epsilon_{k+1}$-sphere centered at $x$ (the direction given by the dashed blue line in \autoref{fig:ddn_illustration}), obtaining $\tilde{x}_{k+1}$. 
Lastly, $\tilde{x}_{k+1}$ is projected onto the feasible region of the input space $\mathcal{X}$. In the case of images normalized to $[0, 1]$, we simply clip the value of each pixel to be inside this range (step 13 of \autoref{alg:ddn}). Besides this step, we can also consider quantizing the image in each iteration, to ensure the attack is a valid image.

It's worth noting that, when reaching a point where the decision boundary is tangent to the $\epsilon_k$-sphere, $g$ will have the same direction as $\delta_{k+1}$. This means that $\delta_{k+1}$ will be projected on the direction of $\delta_k$. Therefore, the norm will oscillate between the two sides of the decision boundary in this direction. Multiplying $\epsilon$ by $1+\gamma$ and $1-\gamma$ will result in a global decrease (on two steps) of the norm by $1-\gamma^2$, leading to a finer search of the best norm.

\section{Attack Evaluation}
\label{sec:experimental_protocal}


\begin{table*}
    \centering
    \resizebox{1.35\columnwidth}{!}{%
    \begin{tabular}{ccl|rrrrr}
    & Attack & Budget & Success & Mean $L_2$ & Median $L_2$ & \#Grads & Run-time (s) \\
    \toprule[1pt]
    \multirow{7}{*}[-0.6em]{\begin{turn}{90}MNIST\end{turn}} & \multirow{3}{*}{C\&W} 
     & 4$\times$25 & 100.0 & 1.7382 & 1.7400 & 100 & 1.7\\
     & & 1$\times$100 & 99.4 & 1.5917 & 1.6405 & 100 & 1.7\\
     & & 9$\times$10\,000 & 100.0 & \textbf{1.3961} & 1.4121 & 54\,007 & 856.8 \\
    \cmidrule{2-8}
    & DeepFool & 100 & 75.4 & 1.9685 & 2.2909 & 98 & - \\
    \cmidrule{2-8}
    &\multirow{3}{*}{DDN} & 100 & 100.0 & 1.4563 & 1.4506 & 100 & 1.5 \\
    & & 300 & 100.0 & 1.4357 & 1.4386 & 300 & 4.5 \\
    & & 1\,000 & 100.0 & 1.4240 & 1.4342 & 1\,000 & 14.9 \\ 
    \midrule[1pt]
    \multirow{7}{*}[-0.6em]{\begin{turn}{90}CIFAR-10\end{turn}} & \multirow{3}{*}{C\&W} 
     & 4$\times$25 & 100.0 & 0.1924 & 0.1541 & 60 & 3.0\\
     & & 1$\times$100 & 99.8 & 0.1728 & 0.1620 & 91 & 4.6\\
     & & 9$\times$10\,000 & 100.0 & 0.1543 & 0.1453 & 36\,009 & 1\,793.2 \\
    \cmidrule{2-8}
    & DeepFool & 100 & 99.7 & 0.1796 & 0.1497 & 25 & - \\
    \cmidrule{2-8}
    & \multirow{3}{*}{DDN} & 100 & 100.0 & 0.1503 & 0.1333 & 100 & 4.7 \\
    & & 300 & 100.0 & 0.1487 & 0.1322 & 300 & 14.2 \\
    & & 1\,000 & 100.0 & \textbf{0.1480} & 0.1317 & 1\,000 & 47.6 \\ 
    \midrule[1pt]
    \multirow{7}{*}[-0.6em]{\begin{turn}{90}ImageNet\end{turn}} & \multirow{3}{*}{C\&W} 
     & 4$\times$25 & 100.0 & 1.5812 & 1.3382 & 63 & 379.3\\
     & & 1$\times$100 & 100.0 & 0.9858 & 0.9587 & 48 & 287.1\\
     & & 9$\times$10\,000& 100.0 & 0.4692 & 0.3980 & 21\,309 & 127\,755.6 \\
    \cmidrule{2-8}
    & DeepFool & 100 & 98.5 & 0.3800 & 0.2655 & 41 & - \\
    \cmidrule{2-8}
    & \multirow{3}{*}{DDN} & 100 & 99.6 & 0.3831 & 0.3227 & 100 & 593.6 \\
    & & 300 & 100.0 & 0.3749 & 0.3210 & 300 & 1\,779.4 \\
    & & 1\,000 & 100.0 & \textbf{0.3617} & 0.3188 & 1\,000 & 5\,933.6 \\ 
    \bottomrule[1pt]
    \end{tabular}
    }
    \caption{Performance of our DDN attack compared to C\&W \cite{carlini_towards_2017} and DeepFool \cite{moosavi2016deepfool} attacks on MNIST, CIFAR-10 and ImageNet in the untargeted scenario.}
    \label{tab:attack_resutls}
    \vspace{-0.5em}
\end{table*}

\begin{table}
    \centering
    \resizebox{\columnwidth}{!}{%
    \begin{tabular}{l|rr|rr}
        &  \multicolumn{2}{c|}{Average case} & \multicolumn{2}{c}{Least Likely} \\
        Attack &  Success & Mean $L_2$ &  Success & Mean $L_2$   \\
        \midrule[1pt]
        C\&W 4$\times$25    & 96.11 & 2.8254 & 69.9 & 5.0090 \\
        C\&W 1$\times$100       & 86.89	& 2.0940 &	31.7 &2.6062 \\
        C\&W 9$\times$10\,000 & 100.00      & 1.9481       &  100.0     &    2.5370    \\
        \midrule[0.5pt]
        DDN 100  & 100.00   & 1.9763 & 100.0   & 2.6008 \\
        DDN 300 & 100.00 & 1.9577 & 100.0 & 2.5503 \\
        DDN 1\,000 & 100.00   & 1.9511 & 100.0   & 2.5348 \\
        \bottomrule[1pt]
    \end{tabular}
    }
    \caption{Comparison of the DDN attack to the C\&W $L_2$ attack on MNIST.}
    \label{tab:targeted_mnist}
\end{table}

\begin{table}
    \centering
    \resizebox{\columnwidth}{!}{%
    \begin{tabular}{l|rr|rr}
        &  \multicolumn{2}{c|}{Average case} & \multicolumn{2}{c}{Least Likely} \\
        Attack &  Success & Mean $L_2$ &  Success & Mean $L_2$   \\
        \midrule[1pt]
        C\&W 4$\times$25 & 99.78 &	0.3247 & 98.7 &	0.5060 \\
        C\&W 1$\times$100  & 99.32 & 0.3104 & 95.8 & 0.4159 \\
        C\&W 9$\times$10\,000 &  100.00 & 0.2798 & 100.0 & 0.3905 \\
        \midrule[0.5pt]
        DDN 100  & 100.00 & 0.2925 & 100.0  & 0.4170  \\
        DDN 300 & 100.00 & 0.2887 & 100.0 & 0.4090 \\
        DDN 1\,000 & 100.00 & 0.2867 & 100.0 & 0.4050 \\
        \bottomrule[1pt]
    \end{tabular}
    }
    \caption{Comparison of the DDN attack to the C\&W $L_2$ attack on CIFAR-10.}
    \label{tab:targeted_cifar}
    \vspace{-1em}
\end{table}

\begin{table}
    \centering
    \resizebox{\columnwidth}{!}{%
    \begin{tabular}{l|rr|rr}
        &\multicolumn{2}{c|}{Average case} & \multicolumn{2}{c}{Least Likely} \\
        Attack &  Success & Mean $L_2$  &  Success & Mean $L_2$   \\
        \midrule[1pt]
        C\&W 4$\times$25    & 99.13	& 4.2826 &	80.6 & 8.7336  \\
        C\&W 1$\times$100   & 96.74& 	1.7718 &	66.2 &	2.2997  \\
        C\&W 9$\times$10\,000 \cite{carlini_towards_2017}  & 100.00 & 0.96 & 100.0 & 2.22 \\
        \midrule[0.5pt]
        DDN 100 &  99.98 & 1.0260 &  99.5 & 1.7074\\
        DDN 300 &  100.00 & 0.9021 &  100.0 & 1.3634\\
        DDN 1\,000 & 100.00 & 0.8444 & 100.0 & 1.2240\\
        \bottomrule[1pt]
    \end{tabular}
    }
    \caption{Comparison of the DDN attack to the C\&W $L_2$ attack on ImageNet. For  C\&W 9$\times$10\,000, we report the results from \cite{carlini_towards_2017}.}
    \label{tab:targeted_imagenet}
    \vspace{-1.2em}
\end{table}

Experiments were conducted on the MNIST, CIFAR-10 and ImageNet datasets, comparing the proposed attack to the state-of-the-art $L_2$ attacks proposed in the literature: DeepFool \cite{moosavi2016deepfool} and C\&W $L_2$ attack \cite{carlini_towards_2017}. We use the same model architectures and hyperparameters for training as in \cite{carlini_towards_2017} for MNIST and CIFAR-10 (see the supplementary material for details). Our base classifiers obtain 99.44\% and 85.51\% accuracy on the test sets of MNIST and CIFAR-10, respectively. For the ImageNet experiments, we use a pre-trained Inception V3 \cite{szegedy_rethinking_2015}, that achieves 22.51\% top-1 error on the validation set. Inception V3 takes images of size 299$\times$299 as input, which are cropped from images of size 342$\times$342.

For experiments with DeepFool \cite{moosavi2016deepfool}, we used the implementation from Foolbox \cite{rauber2017foolbox}, with a budget of 100 iterations. For the experiments with C\&W, we ported the attack (originally implemented on TensorFlow) on PyTorch to evaluate the models in the frameworks in which they were trained. We use the same hyperparameters from \cite{carlini_towards_2017}: 9 search steps on C with an initial constant of $0.01$, with 10\,000 iterations for each search step (with early stopping) - we refer to this scenario as C\&W 9$\times$10\,000 in the tables.
As we are interested in obtaining attacks that require few iterations, we also report experiments in a scenario where the number of iterations is limited to 100. We consider a scenario of running 100 steps with a fixed $C$ (1$\times$100), and a scenario of running 4 search steps on $C$, of 25 iterations each (4$\times$25). Since the hyperparameters proposed in \cite{carlini_towards_2017} were tuned for a larger number of iterations and search steps, we performed a grid search for each dataset, using learning rates in the range [0.01, 0.05, 0.1, 0.5, 1], and $C$ in the range [0.001, 0.01, 0.1, 1, 10, 100, 1\,000]. We report the results for C\&W with the hyperparameters that achieve best Median $L_2$. Selected parameters are listed in the supplementary material.

For the experiments using DDN, we ran attacks with budgets of 100, 300 and 1\,000 iterations, in all cases, using $\epsilon_0 = 1$ and $\gamma = 0.05$. The initial step size $\alpha = 1$, was reduced with cosine annealing to $0.01$ in the last iteration. The choice of $\gamma$ is based on the encoding of images. For any correctly classified image, the smallest possible perturbation consists in changing one pixel by $1/255$ (for images encoded in 8 bit values), corresponding to a norm of $1/255$. Since we perform quantization, the values are rounded, meaning that the algorithm must be able to achieve a norm lower than $1.5/255 = 3/510$. When using $K$ steps, this imposes:
\begin{equation}
    \epsilon_0(1-\gamma)^K < \frac{3}{510} \Rightarrow \gamma > 1 - \bigg(\frac{3}{510\, \epsilon_0}\bigg)^{\frac{1}{K}}
\end{equation}
Using $\epsilon_0 = 1$ and $K=100$ yields $\gamma \simeq 0.05$. Therefore, if there exists an adversarial example with smallest perturbation, the algorithm may find it in a fixed number of steps.

For the results with DDN, we consider quantized images (to 256 levels). The quantization step is included in each iteration (see step 13 of \autoref{alg:ddn}). All results reported in the paper consider images in the $[0, 1]$ range.

Two sets of experiments were conducted: untargeted attacks and targeted attacks.  As in \cite{carlini_towards_2017}, we generated attacks on the first 1\,000 images of the test set for MNIST and CIFAR-10, while for ImageNet we randomly chose 1\,000 images from the validation set that are correctly classified.  For the untargeted attacks, we report the success rate of the attack (percentage of samples for which an attack was found), the mean $L_2$ norm of the adversarial noise (for successful attacks), and the median $L_2$ norm over all attacks while considering unsuccessful attacks as worst-case adversarial (distance to a uniform gray image, as in \cite{brendel2018adversarial}). We also report the average number (for batch execution) of gradient computations and the total run-times (in seconds) on a NVIDIA GTX 1080 Ti with 11GB of memory. We did not report run-times for the DeepFool attack, since the implementation from foolbox generates adversarial examples one-by-one and is executed on CPU, leading to unrepresentative run-times. Attacks on MNIST and CIFAR-10 have been executed in a single batch of 1\,000 samples, whereas attacks on ImageNet have been executed in 20 batches of 50 samples.

For the targeted attacks, following the protocol from \cite{carlini_towards_2017}, we generate attacks against all possible classes on MNIST and CIFAR-10 (9 attacks per image), and against 100 randomly chosen classes for ImageNet (10\% of the number of classes). Therefore, in each targeted attack experiment, we run 9\,000 attacks on MNIST and CIFAR-10, and 100\,000 attacks on ImageNet. Results are reported for two scenarios: 1) average over all attacks; 2) average performance when choosing the least likely class (i.e. choosing the worst attack performance over all target classes, for each image). The reported $L_2$ norms are, as in the untargeted scenario, the means over successful attacks.

\autoref{tab:attack_resutls} reports the results of DDN compared to the C\&W $L_2$ and DeepFool attacks on the MNIST, CIFAR-10 and ImageNet datasets. 
For the MNIST and CIFAR-10 datasets, results with DDN are comparable to the state-of-the-art. DDN obtains slightly worse $L_2$ norms on the MNIST dataset (when compared to the C\&W 9$\times$10\,000), however, our attack is able to get within 5\% of the norm found by C\&W in only 100 iterations compared to the 54\,007 iterations required for the C\&W $L_2$ attack. When the C\&W attack is restricted to use a maximum of 100 iterations, it always performed worse than DDN with 100 iterations. On the ImageNet dataset, our attack obtains better Mean $L_2$ norms than both other attacks. The DDN attack needs 300 iterations to reach 100\% success rate. DeepFool obtains close results but fails to reach 100\% success rate. It is also worth noting that DeepFool seems to performs worse against adversarially trained models (discussed in \autoref{sec:defense_evaluation}). 
Supplementary material reports curves of the perturbation size against accuracy of the models for the three attacks.

Tables \ref{tab:targeted_mnist}, \ref{tab:targeted_cifar} and \ref{tab:targeted_imagenet} present the results on targeted attacks on the MNIST, CIFAR-10 and ImageNet datasets, respectively. For the MNIST and CIFAR-10 datasets, DDN yields similar performance compared to the C\&W attack with 9$\times$10\,000 iterations, and always perform better than the C\&W attack when it is restricted to 100 iterations (we re-iterate that the hyperparameters for the C\&W attack were tuned for each dataset, while the hyperparameters for DDN are fixed for all experiments). On the ImageNet dataset, DDN run with 100 iterations obtains superior performance than C\&W. For all datasets, with the scenario restricted to 100 iterations, the C\&W algorithm has a noticeable drop in success rate for finding adversarial examples to the least likely class.

\section{Adversarial Training with DDN}
\label{sec:defense}
Since the DDN attack can produce adversarial examples in relatively few iterations, it can be used for adversarial training. For this, we consider the following loss function:

\begin{equation}
    \tilde{J}(x, y, \theta) = J(\tilde{x}, y, \theta)
    \label{eq:our_defense}
\end{equation}

\noindent where $\tilde{x}$ is an adversarial example produced by the DDN algorithm, that is projected to an $\epsilon$-ball around $x$, such that the classifier is trained with adversarial examples with a maximum norm of $\epsilon$. 
It is worth making a parallel of this approach with the Madry defense \cite{madry_towards_2017} where, in each iteration, the loss of the worst-case adversarial (see \autoref{eq:worst_adv}) in an $\epsilon$-ball around the original sample $x$ is used for optimization. In our proposed adversarial training procedure, we optimize the loss of the closest adversarial example (see \autoref{eq:closer_adv}). The intuition of this defense is to push the decision boundary away from $x$ in each iteration. We do note that this method does not have the theoretical guarantees of the Madry defense. However, since in practice the Madry defense uses approximations (when searching for the global maximum of the loss around $x$), we argue that both methods deserve empirical comparison.

\section{Defense Evaluation}
\label{sec:defense_evaluation}

We trained models using the same architectures as \cite{carlini_towards_2017} for MNIST, and a Wide ResNet (WRN) 28-10 \cite{zagoruyko2016wide} for CIFAR-10 (similar to \cite{madry_towards_2017} where they use a WRN 34-10). As described in \autoref{sec:defense}, we augment the training images with adversarial perturbations. For each training step, we run the DDN attack with a budget of 100 iterations, and limit the norm of the perturbation to a maximum $\epsilon = 2.4$ on the MNIST experiments, and $\epsilon = 1$ for the CIFAR-10 experiments. For MNIST, we train the model for 30 epochs with a learning rate of $0.01$ and then for 20 epochs with a learning rate of $0.001$. To reduce the training time with CIFAR-10, we first train the model on original images for 200 epochs using the hyperparameters from \cite{zagoruyko2016wide}. Then, we continue training for 30 more epochs using \autoref{eq:our_defense}, keeping the same final learning rate of $0.0008$. Our robust MNIST model has a test accuracy of 99.01\% on the clean samples, while the Madry model has an accuracy of 98.53\%. On CIFAR-10, our model reaches a test accuracy of 89.0\% while the model by Madry \etal  obtains 87.3\%.

\begin{table}
    \centering
    \resizebox{\columnwidth}{!}{%
    \begin{tabular}{clrrrr}
    Defense & Attack & \begin{tabular}{c}
         Attack \\
         Success 
    \end{tabular} & Mean $L_2$ & Median $L_2$ & \begin{tabular}{c}
         Model\\Accuracy \\
         at $\epsilon \leq 1.5$ 
    \end{tabular} \\
    \toprule[1pt]
    \multirow{4}{*}[-2pt]{Baseline}
    & C\&W 9$\times$10\,000 & 100.0 & 1.3961 & 1.4121 & 42.1 \\
    & DeepFool 100 & 75.4 & 1.9685 & 2.2909 & 81.8 \\ 
    & DDN 1\,000 & 100.0 & 1.4240 & 1.4342 & 45.2 \\
    \cmidrule[0.1pt]{2-6}
    & \textbf{All} & 100.0 & 1.3778 & 1.3946 & 40.8 \\
    \midrule[0.5pt]
    \multirow{4}{*}[-4pt]{\begin{tabular}{c}Madry \\\etal\end{tabular}}
    & C\&W 9$\times$10\,000 & 100.0 & 2.0813 & 2.1071 & 73.0 \\
    & DeepFool 100 & 91.6 & 4.9585 & 5.2946 & 93.1 \\ 
    & DDN 1\,000 & 99.6 & 1.8436 & 1.8994 & 69.9 \\
    \cmidrule[0.1pt]{2-6}
    & \textbf{All} & 100.0 & 1.6917 & 1.8307 & 67.3 \\
    \midrule[0.5pt]
    \multirow{4}{*}[-2pt]{Ours}
    & C\&W 9$\times$10\,000 & 100.0 & 2.5181 & 2.6146 & 88.0 \\
    & DeepFool 100 & 94.3 & 3.9449 & 4.1754 & 92.7 \\
    & DDN 1\,000 & 100.0 & 2.4874 & 2.5781 & 87.6 \\
    \cmidrule[0.1pt]{2-6}
    & \textbf{All} & 100.0 & \textbf{2.4497} & 2.5538 & \textbf{87.2} \\
    \bottomrule[1pt]
    \end{tabular}}
    \caption{Evaluation of the robustness of our adversarial training on MNIST against the Madry defense.}
    \label{tab:mnist_defense}
    \vspace{-1em}
\end{table}

\begin{table}
    \centering
    \resizebox{\columnwidth}{!}{%
    \begin{tabular}{clrrrr}
    Defense & Attack & \begin{tabular}{c}
         Attack \\
         Success 
    \end{tabular} & Mean $L_2$ & Median $L_2$ & \begin{tabular}{c}
         Model\\Accuracy \\
         at $\epsilon \leq 0.5$ 
    \end{tabular} \\
    \toprule[1pt]
    \multirow{4}{*}[-3pt]{\begin{tabular}{c}Baseline\\WRN 28-10\end{tabular}} 
    & C\&W 9$\times$10\,000 & 100.0 & 0.1343 & 0.1273 & 0.2 \\
    & DeepFool 100 & 99.3 & 0.5085 & 0.4241 & 38.3 \\ 
    & DDN 1\,000 & 100.0 & 0.1430 & 0.1370 & 0.1 \\
    \cmidrule[0.1pt]{2-6}
    & \textbf{All} & 100.0 & 0.1282 & 0.1222 & 0.1 \\
    \midrule[0.5pt]
    \multirow{4}{*}[-3pt]{\begin{tabular}{c}Madry\\\etal\\WRN 34-10\end{tabular}}
    & C\&W 9$\times$10\,000 & 100.0 & 0.6912 & 0.6050 & 57.1 \\
    & DeepFool 100 & 95.6 & 1.4856 & 0.9576 & 64.7 \\ 
    & DDN 1\,000 & 100.0 & 0.6732 & 0.5876 & 56.9 \\
    \cmidrule[0.1pt]{2-6}
    & \textbf{All} & 100.0 & 0.6601 & 0.5804 & 56.1 \\
    \midrule[0.5pt]
    \multirow{4}{*}[-3pt]{\begin{tabular}{c}Ours\\WRN 28-10\end{tabular}} 
    & C\&W 9$\times$10\,000 & 100.0 & 0.8860 & 0.8254 & 67.9 \\
    & DeepFool 100 & 99.7 & 1.5298 & 1.1163 & 69.9 \\
    & DDN 1\,000 & 100.0 & 0.8688 & 0.8177 & 68.0 \\
    \cmidrule[0.1pt]{2-6}
    & \textbf{All} & 100.0 & \textbf{0.8597} & 0.8151 & \textbf{67.6} \\
    \bottomrule[1pt]
    \end{tabular}}
    \caption{Evaluation of the robustness of our adversarial training on CIFAR-10 against the Madry defense.}
    \label{tab:cifar10_defense}
\end{table}

We evaluate the adversarial robustness of the models using three untargeted attacks: Carlini 9$\times$10\,000, DeepFool 100 and DDN 1\,000. For each sample, we consider the smallest adversarial perturbation produced by the three attacks and report it in the ``\textbf{All}'' row. 
 Tables \ref{tab:mnist_defense} and \ref{tab:cifar10_defense} report the results of this evaluation with a comparison to the defense of Madry \etal \cite{madry_towards_2017}\footnote{Models taken from \url{https://github.com/MadryLab}} and the baseline (without adversarial training) for CIFAR-10. For MNIST, the baseline corresponds to the model used in \autoref{sec:experimental_protocal}. 
We observe that for attacks with unbounded norm, the attacks can successfully generate adversarial examples almost 100\% of the time. However, an increased $L_2$ norm is required to generate attacks against the model trained with DDN. 

\begin{figure}
    \centering
    \includegraphics[width=\columnwidth]{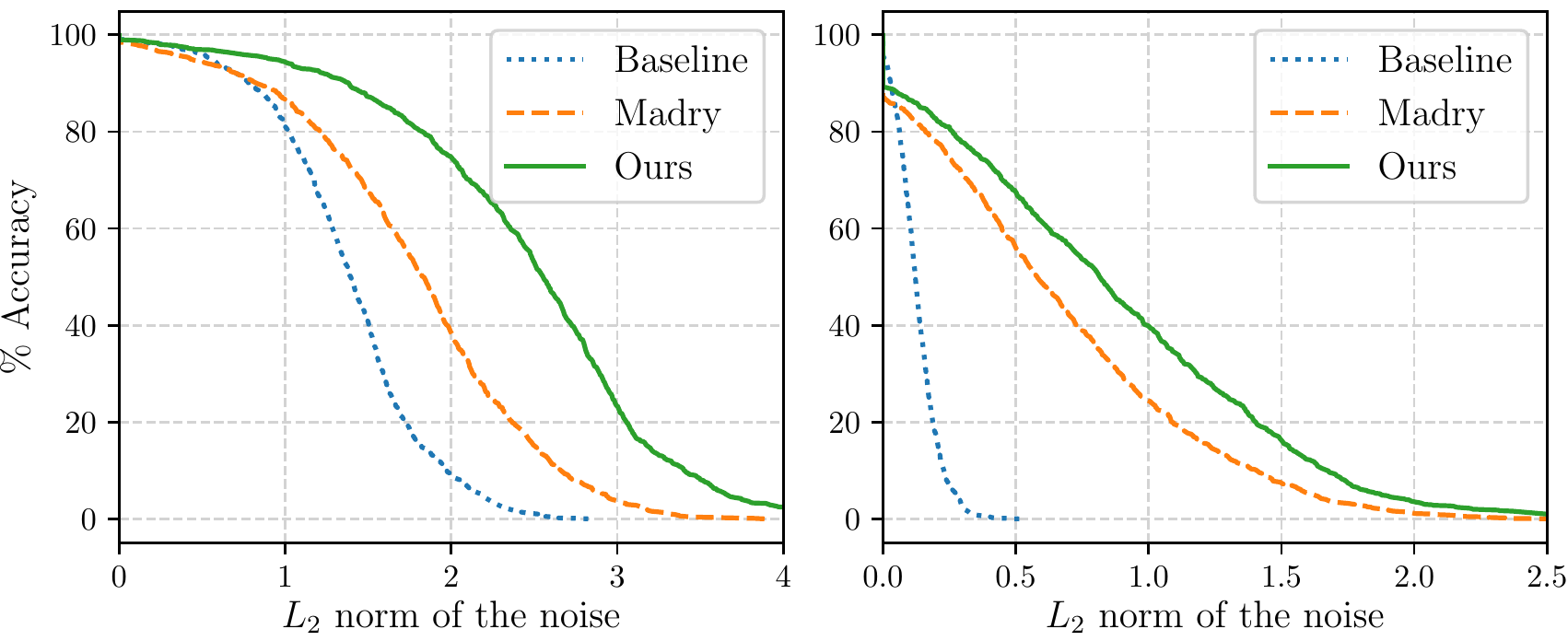}
    \caption{Models robustness on MNIST (left) and CIFAR-10 (right): impact on accuracy as we increase the maximum perturbation $\epsilon$.}
    \label{fig:models_robustness_vs_noise}
    \vspace{-1em}
\end{figure}

\autoref{fig:models_robustness_vs_noise} shows the robustness of the MNIST and CIFAR-10 models respectively for different attacks with increasing maximum $L_2$ norm. These figures can be interpreted as the expected accuracy of the systems in a scenario where the adversary is constrained to make changes with norm $L_2 \le \epsilon$. For instance on MNIST, if the attacker is limited to a maximum norm of $\epsilon = 1.5$, the baseline performance decreases to 40.8\%; Madry to 67.3\% and our defense to 87.2\%. At $\epsilon = 2.0$, baseline performance decreases to 9.2\%, Madry to 38.6\% and our defense to 74.8\%. On CIFAR-10, if the attacker is limited to a maximum norm of $\epsilon = 0.5$, the baseline performance decreases to 0.1\%; Madry to 56.1\% and our defense to 67.6\%. At $\epsilon = 1.0$, baseline performance decreases to 0\%, Madry to 24.4\% and our defense to 39.9\%. For both datasets, the model trained with DDN outperforms the model trained with the Madry defense for all values of $\epsilon$.

\begin{figure}
    \centering
    \includegraphics[width=\columnwidth]{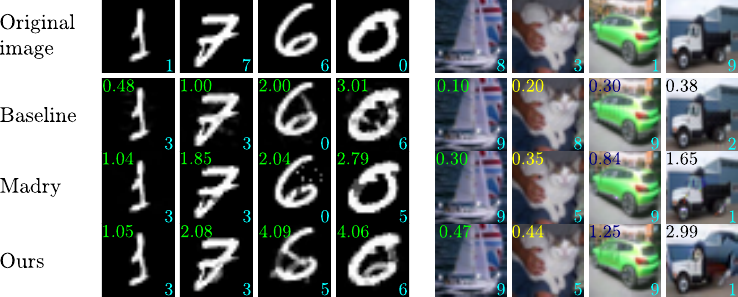}
    \caption{Adversarial examples with varied levels of noise $\delta$ against three models: baseline, Madry defense \cite{madry_towards_2017} and our defense. Text on top-left of each image indicate $\norm{\delta}_2$; text on bottom-right indicates the predicted class\protect\footnotemark.}
    \label{fig:illustration_adv}
    \vspace{-1.2em}
\end{figure}

\autoref{fig:illustration_adv} shows adversarial examples produced by the DDN 1\,000 attack for different models on MNIST and CIFAR-10. On MNIST, adversarial examples for the baseline are not meaningful (the still visually belong to the original class), whereas some adversarial examples obtained for the adversarially trained model (DDN) actually change classes (bottom right: 0 changes to 6). For all models, there are still some adversarial examples that are very close to the original images (first column). On CIFAR-10, while the adversarially trained models require higher norms for the attacks, most adversarial examples still perceptually resemble the original images. In few cases (bottom-right example for CIFAR-10), it could cause a confusion: it can appear as changing to class 1 - a (cropped) automobile facing right.

\footnotetext{For CIFAR-10: 1: automobile, 2: bird, 3: cat, 5: dog, 8: ship, 9: truck.}

\section{Conclusion}
\label{sec:conclusion}

We presented the \emph{Decoupled Direction and Norm} attack, which obtains comparable results with the state-of-the-art for $L_2$ norm adversarial perturbations, but in much fewer iterations. Our attack allows for faster evaluation of the robustness of differentiable models, and enables a novel adversarial training, where, at each iteration, we train with examples close to the decision boundary. 
Our experiments with MNIST and CIFAR-10 show state-of-the-art robustness against $L_2$-based attacks in a white-box scenario. 

The methods presented in this paper were used in NIPS 2018 Adversarial Vision Challenge \cite{brendel2018adversarial}, ranking first in untargeted attacks, and third in targeted attacks and robust models (both attacks and defense in a black-box scenario).
These results highlight the effectiveness of the defense mechanism, and suggest that attacks using adversarially-trained surrogate models can be effective in black-box scenarios, which is a promising future direction.

\section*{Acknowledgements}
We thank Marco Pedersoli and Christian Desrosiers for their insightful feedback. This research was supported by the Fonds de recherche du Québec - Nature et technologies,  Natural Sciences and Engineering Research Council of Canada, and CNPq grant \#206318/2014-6.

{\small
\bibliographystyle{ieee}
\bibliography{egbib}
}

\clearpage
\appendix


\section*{\begin{Large}Supplementary material\end{Large}}

\section{Model architectures}

\newcommand{\ra}[1]{\renewcommand{\arraystretch}{#1}}

\autoref{tab:cnns_attack} lists the architectures of the CNNs used in the Attack Evaluation - we used the same architecture as in \cite{carlini_towards_2017} for a fair comparison against the C\&W and DeepFool attacks. \autoref{tab:cnns_defense} lists the architecture used in the robust model (defense) trained on CIFAR-10. We used a Wide ResNet with $28$ layers and widening factor of $10$ (WRN-28-10). The residual blocks used are the ``basic block" \cite{he_deep_2015, zagoruyko2016wide}, with stride $1$ for the first group and stride $2$ for the second an third groups. This architecture is slightly different from the one used by Madry \etal \cite{madry_towards_2017}, where they use a modified version of Wide ResNet with $5$ residual blocks instead of $4$ in each group, and without convolutions in the residual connections (when the shape of the output changes, e.g. with stride=$2$). 

\begin{table}[b] \centering
\resizebox{\columnwidth}{!}{%
\ra{1.2}
\begin{tabular}{lll}
\toprule
Layer Type & MNIST Model & CIFAR-10 Model \\ 
\midrule

Convolution + ReLU & $3\times 3 \times 32$ & $3\times 3 \times 64$\\
Convolution + ReLU & $3\times 3 \times 32$ & $3\times 3 \times 64$\\
Max Pooling & $2 \times 2$& $2 \times 2$\\
Convolution + ReLU & $3\times 3 \times 64$ & $3\times 3 \times 128$\\
Convolution + ReLU & $3\times 3 \times 64$ & $3\times 3 \times 128$\\
Max Pooling & $2 \times 2$& $2 \times 2$\\
Fully Connected + ReLU & 200 & 256 \\
Fully Connected + ReLU & 200 & 256 \\
Fully Connected + Softmax & 10 & 10 \\

\bottomrule
\end{tabular}
}
\caption{CNN architectures used for the Attack Evaluation}
\label{tab:cnns_attack}
\end{table}

\section{Hyperparameters selected for the C\&W \\attack}

\begin{table}[t] \centering
\resizebox{0.8\columnwidth}{!}{%
\ra{1.2}
\begin{tabular}{lc} 
\toprule
Layer Type & Size \\ 
\midrule
Convolution & $3 \times 3 \times 16$ \\ \vspace{0.5em}
Residual Block & 
$\begin{bmatrix}
    3 \times 3, 160 \\
    3 \times 3, 160 \\
\end{bmatrix} \times 4 $ \\ \vspace{0.5em}
Residual Block & 
$\begin{bmatrix}
    3 \times 3, 320 \\
    3 \times 3, 320 \\
\end{bmatrix} \times 4 $ \\ \vspace{0.5em}
Residual Block & 
$\begin{bmatrix}
    3 \times 3, 640 \\
    3 \times 3, 640 \\
\end{bmatrix} \times 4 $ \\
Batch Normalization + ReLU & - \\
Average Pooling & $ 8 \times 8$ \\
Fully Connected + Softmax & 10  \\
\bottomrule
\end{tabular}
}
\caption{CIFAR-10 architecture used for the Defense evaluation}
\label{tab:cnns_defense}
\end{table}

We considered a scenario of running the C\&W attack with 100 steps and a fixed $C$ (1$\times$100), and a scenario of running 4 search steps on $C$, of 25 iterations each (4$\times$25). Since the hyperparameters proposed in \cite{carlini_towards_2017} were tuned for a larger number of iterations and search steps, we performed a grid search for each dataset, using learning rates in the range [0.01, 0.05, 0.1, 0.5, 1], and $C$ in the range [0.001, 0.01, 0.1, 1, 10, 100, 1\,000]. We selected the hyperparameters that resulted in targeted attacks with lowest Median $L_2$ for each dataset. \autoref{tab:carlini_hyperparams} lists the hyperparameters found through this search procedure.

\begin{table}    \centering
\resizebox{0.7\columnwidth}{!}{%
    \begin{tabular}{lll}
    \toprule
    Dataset & \# Iterations & Parameters \\
    \midrule
        MNIST & $1 \times 100$ &$\alpha=0.1$, $C=1$ \\
        MNIST & $4 \times 25$ &$\alpha=0.5$, $C=1$ \\
        CIFAR-10 & $1 \times 100$ &$\alpha=0.01$, $C=0.1$ \\
        CIFAR-10 & $4 \times 25$ &$\alpha=0.01$, $C=0.1$ \\
        ImageNet & $1 \times 100$ &$\alpha=0.01$, $C=1$ \\
        ImageNet & $4 \times 25$ &$\alpha=0.01$, $C=10$ \\
    \bottomrule
    \end{tabular}
    }
    \caption{Hyperparameters used for the C\&W attack when restricted to 100 iterations.}
    \label{tab:carlini_hyperparams}
\end{table}

\section{Examples of adversarial images}

\begin{figure*}
    \centering
    \subfloat[Baseline (without adversarial training)]{
    \includegraphics[width=0.45\textwidth]{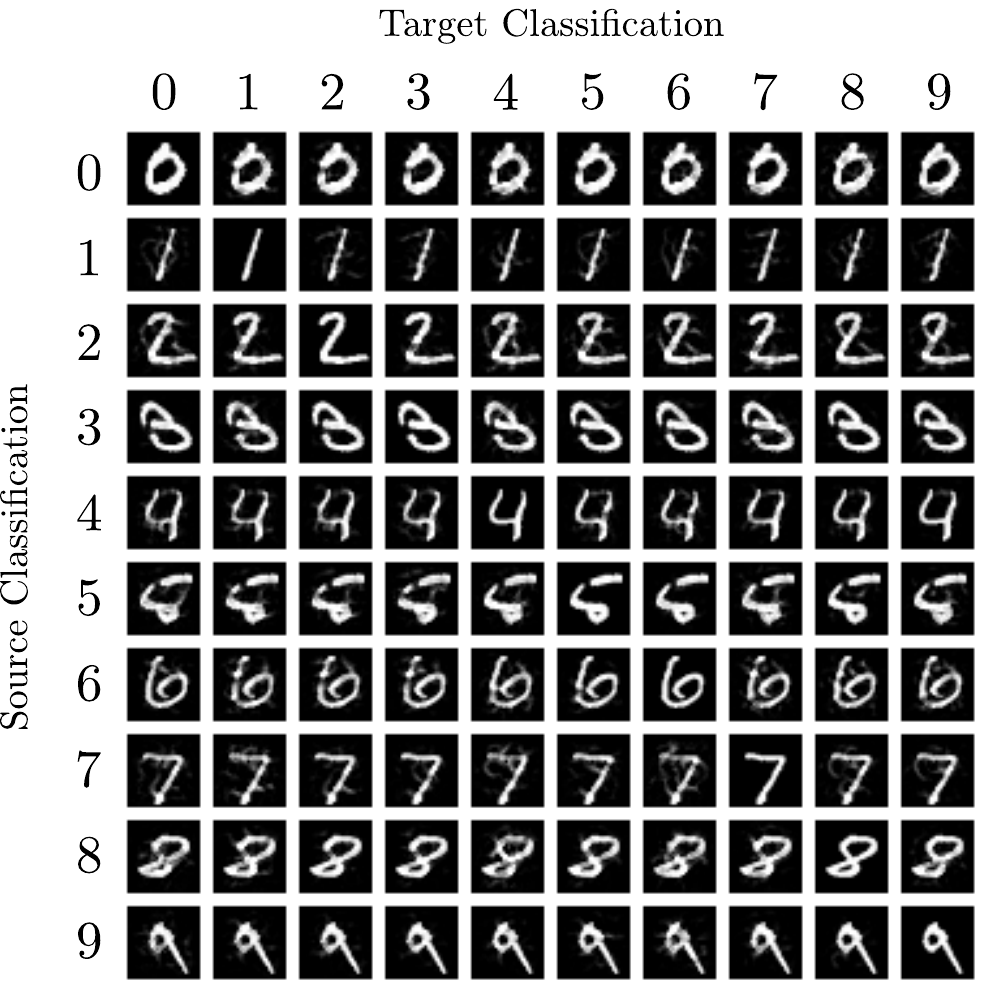}
    }
    \qquad
    \subfloat[Adversarially trained]{\includegraphics[width=0.45\textwidth]
    {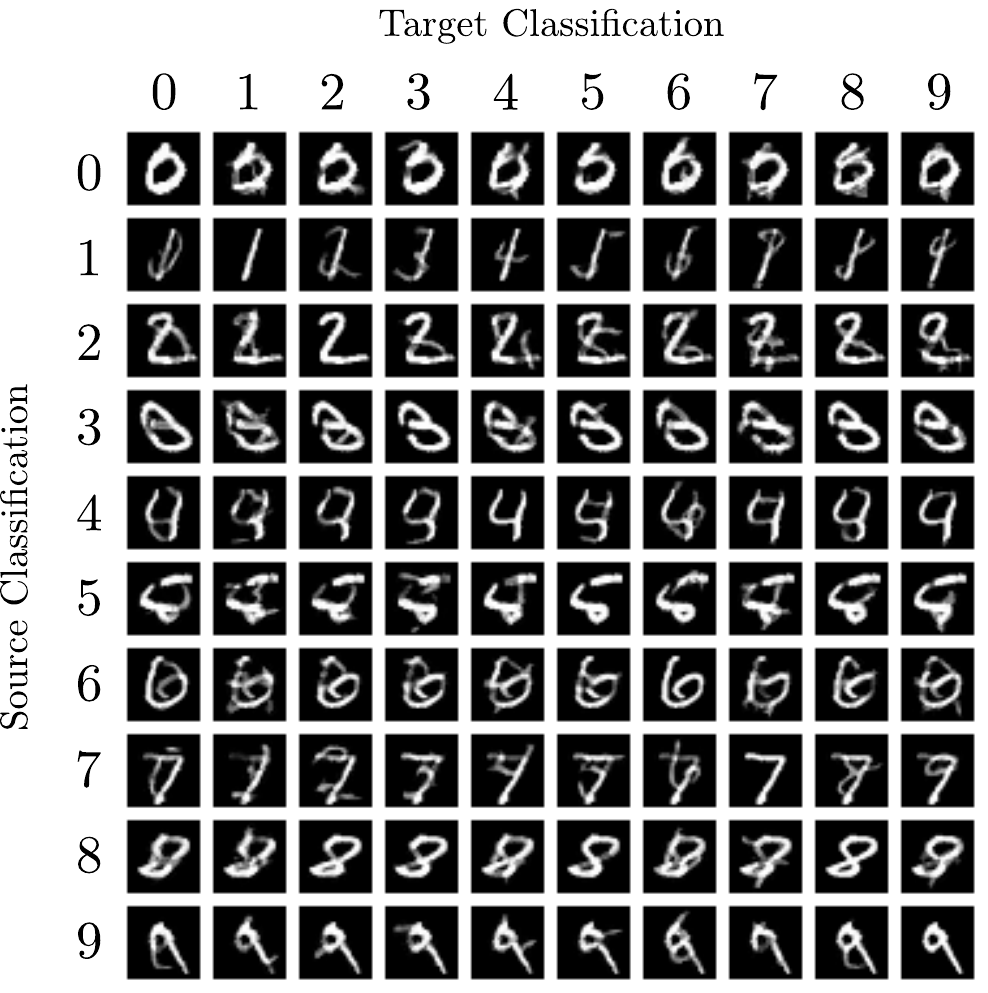}}
    \caption{Adversarial examples obtained using the C\&W $L_2$ attack on two models: (a) Baseline, (b) model adversarially trained with our attack.}
    \label{fig:grid_of_attacks}
\end{figure*}

\autoref{fig:grid_of_attacks} plots a grid of attacks (obtained with the C\&W attack) against the first 10 examples in the MNIST dataset. The rows indicate the source classification (label), and the columns indicate the target class used to generate the attack (images on the diagonal are the original samples). We can see that in the adversarially trained model, the attacks need to introduce much larger changes to the samples in order to make them adversarial, and some of the adversarial samples visually resemble another class.

\begin{figure*}
    \centering
    \includegraphics[width=\textwidth]{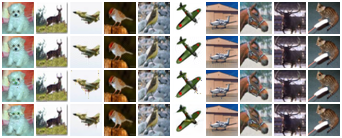}
    \caption{Randomly chosen adversarial examples on CIFAR-10 for three models. \textbf{Top row}: original images; \textbf{second row}: attacks against the baseline; \textbf{third row}: attacks against the Madry defense.}
    \label{fig:cifar_random_picked}
\end{figure*}

\begin{figure*}
    \centering
    \includegraphics[width=\textwidth]{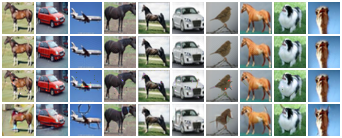}
    \caption{Cherry-picked adversarial examples on CIFAR-10 for three models. \textbf{Top row}: original images; \textbf{second row}: attacks against the baseline; \textbf{third row}: attacks against the Madry defense; \textbf{bottom row}: attacks against the proposed defense. Predicted labels for the last row are, from left to right: dog, ship, deer, dog, dog, truck, horse, dog, cat, cat.}
    \label{fig:cifar_cherry_picked}
\end{figure*}

\autoref{fig:cifar_random_picked} shows randomly-selected adversarial examples for the CIFAR-10 dataset, comparing the baseline model (WRN 28-10), the Madry defense and our proposed defense. For each image and model, we ran three attacks (DDN 1\,000, C\&W 9$\times$10\,000, DeepFool 100), and present the adversarial example with minimum $L_2$ perturbation among them. \autoref{fig:cifar_cherry_picked} shows cherry-picked adversarial examples on CIFAR-10, that visually resemble another class, when attacking the proposed defense. We see that on the average case (randomly-selected), adversarial examples against the defenses still require low amounts of noise (perceptually) to induce misclassification. On the other hand, we notice that on adversarially trained models, some examples do require a much larger change on the image, making it effectively resemble another class.

\section{Attack performance curves}

\begin{figure*}
    \centering
    \subfloat[MNIST / Baseline model.]
    {\includegraphics[width=0.75\columnwidth]{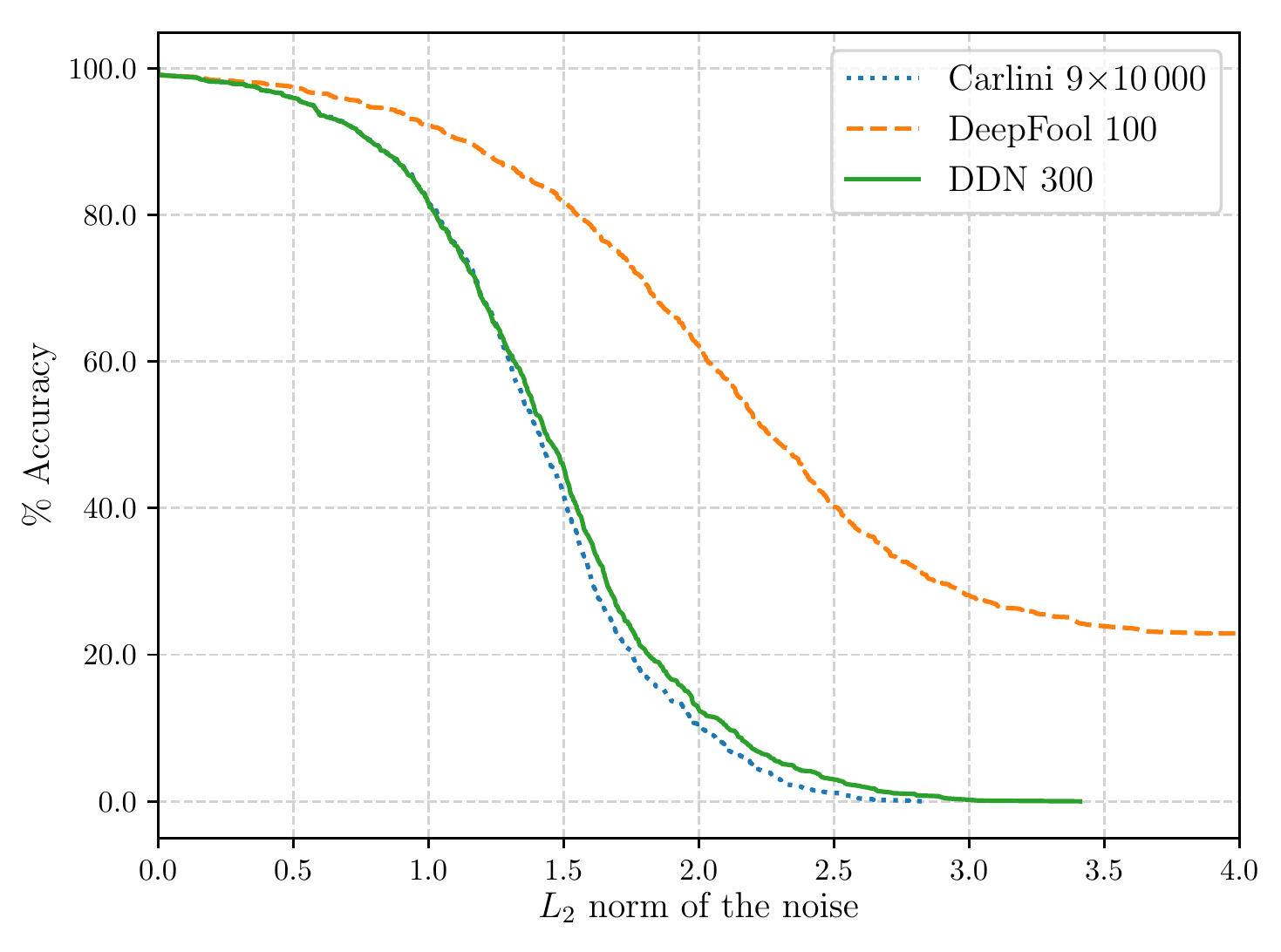}}
    \subfloat[MNIST / Madry defense.]
    {\includegraphics[width=0.75\columnwidth]{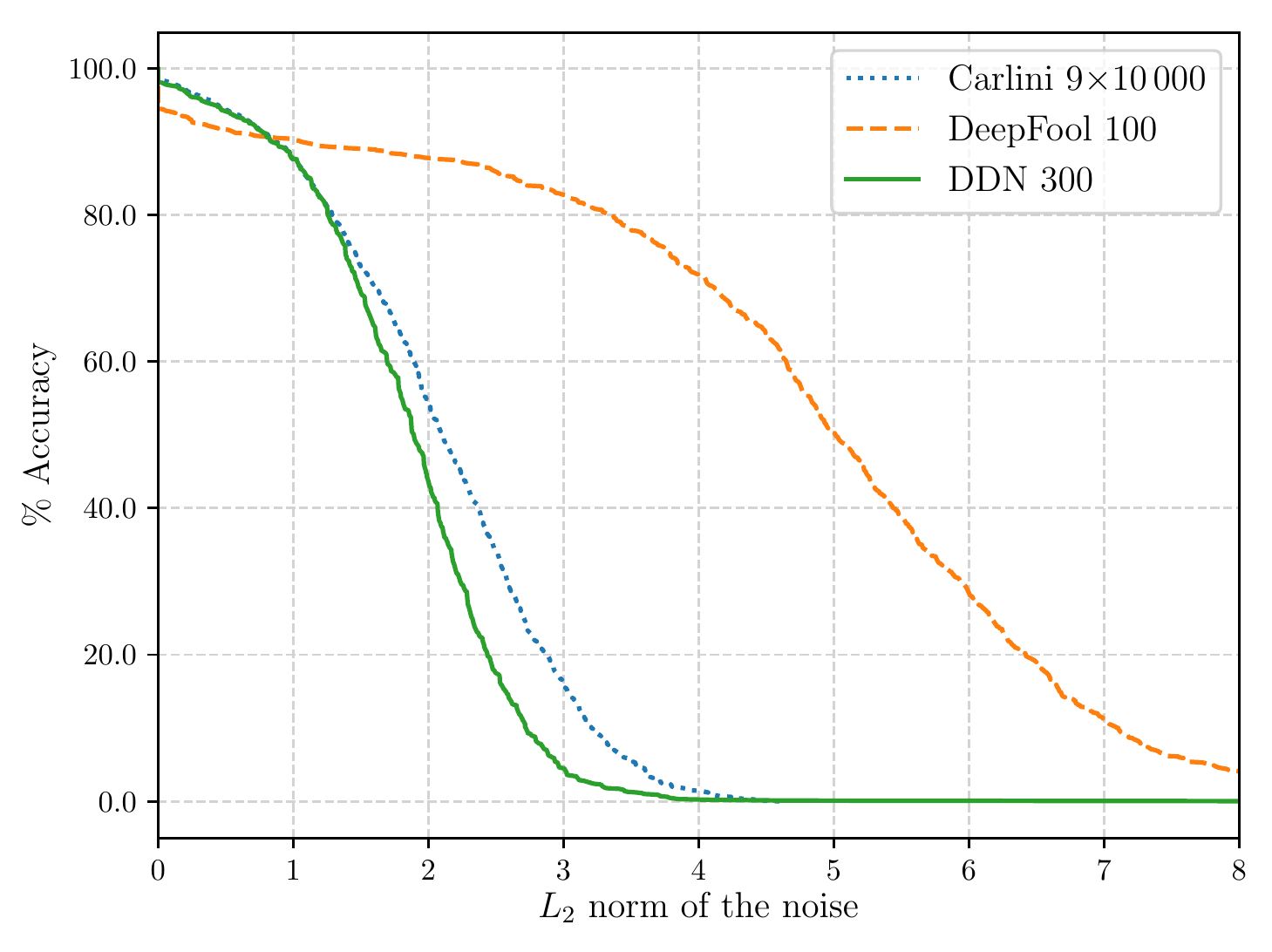}}
    
    \subfloat[MNIST / Our Defense]
    {\includegraphics[width=0.75\columnwidth]{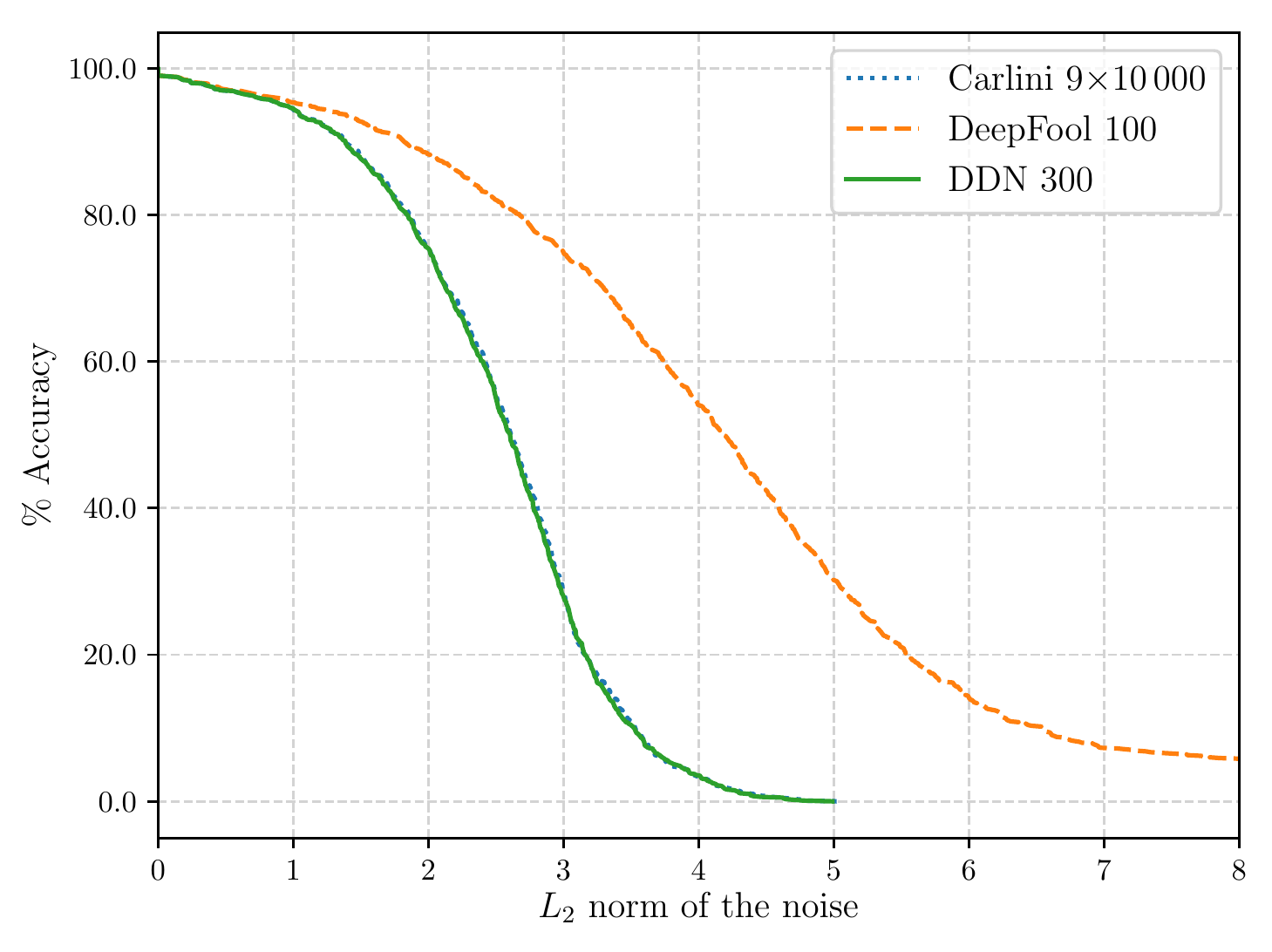}}
    \subfloat[ImageNet / Inception V3.]
    {\includegraphics[width=0.75\columnwidth]{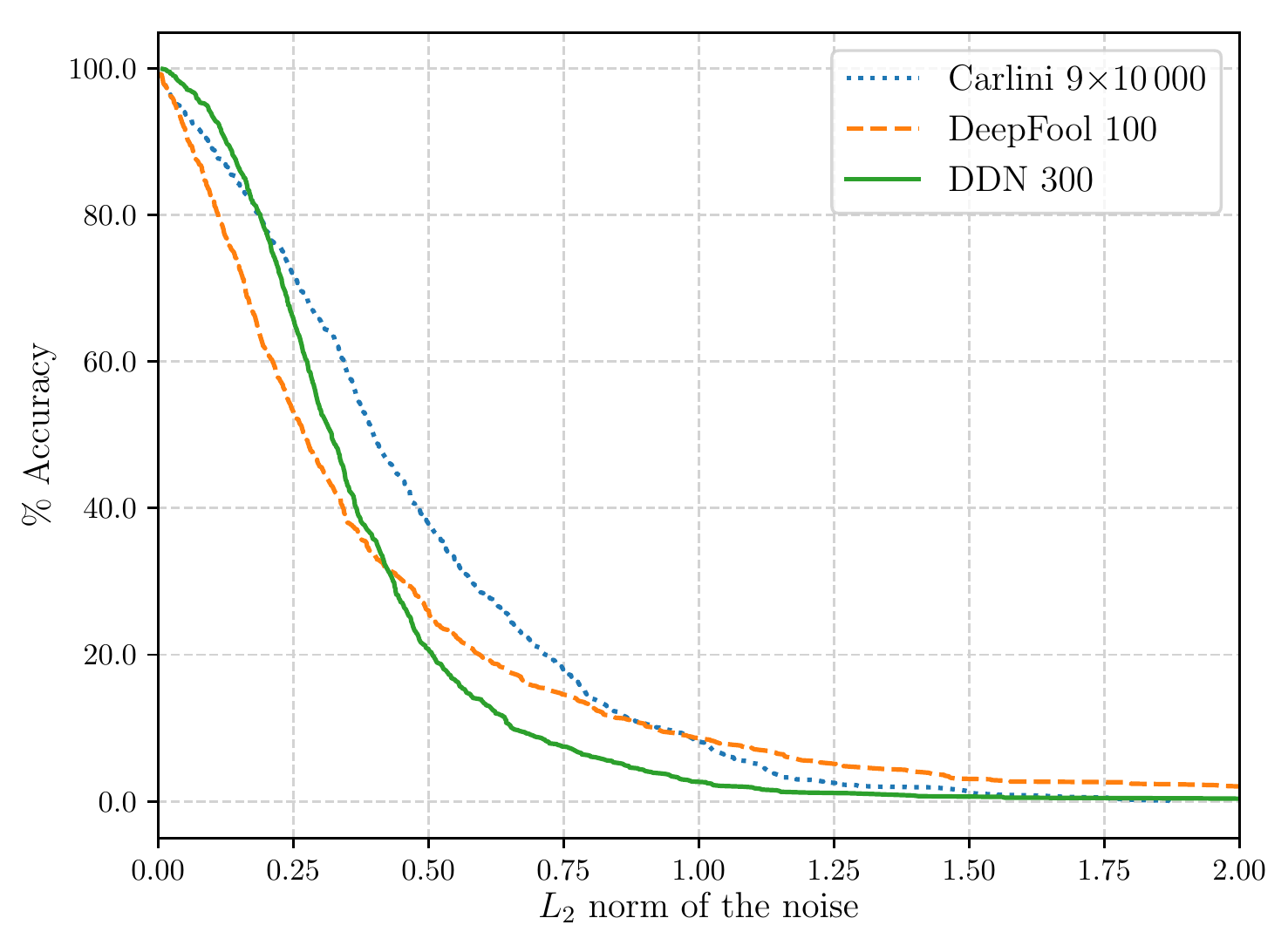}}
    
    \subfloat[CIFAR-10 / Baseline model.]
    {\includegraphics[width=0.75\columnwidth]{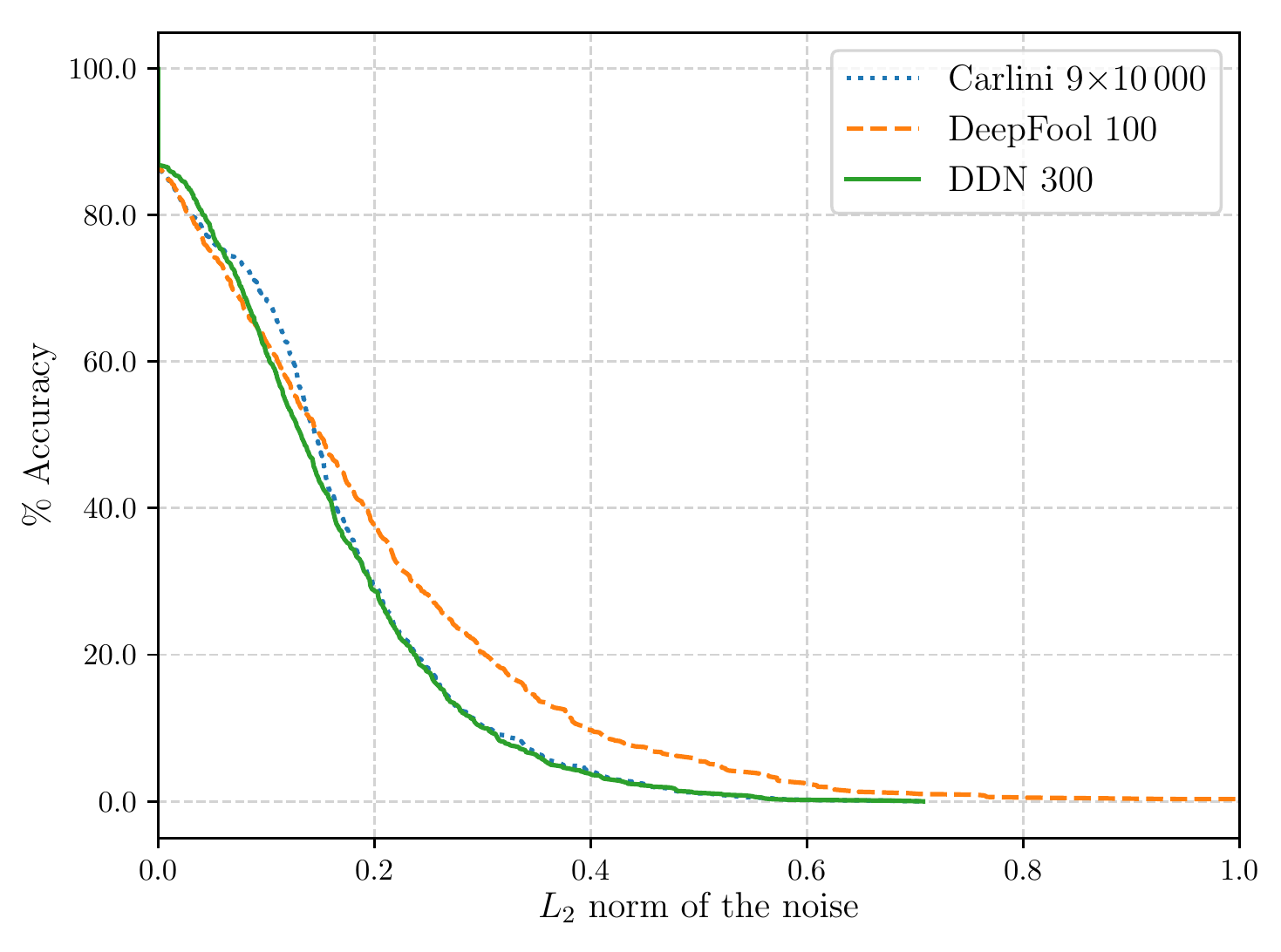}}
    \subfloat[CIFAR-10 / Baseline WRN 28-10.]
    {\includegraphics[width=0.75\columnwidth]{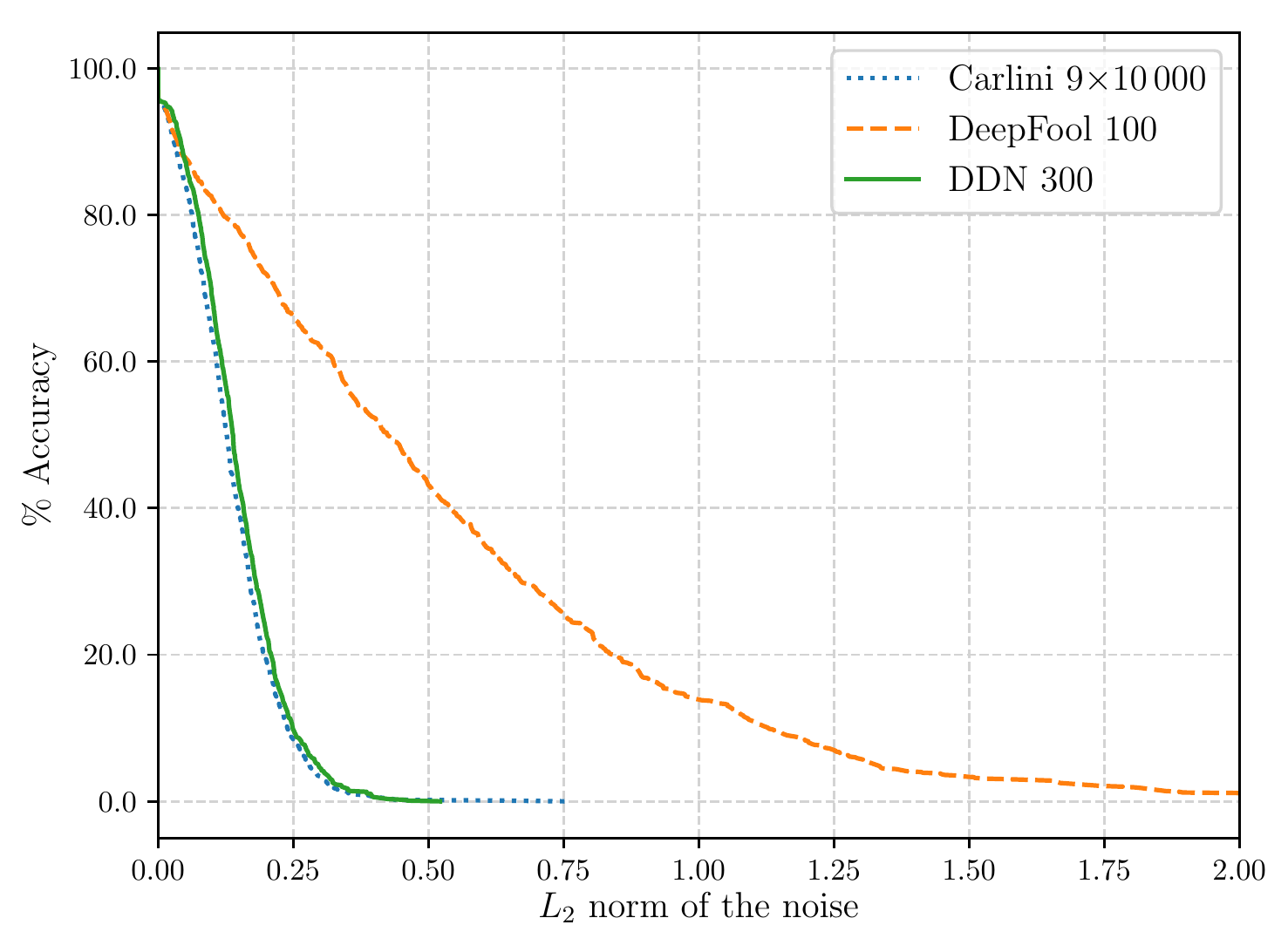}}
    
    \subfloat[CIFAR-10 / Madry defense.]
    {\includegraphics[width=0.75\columnwidth]{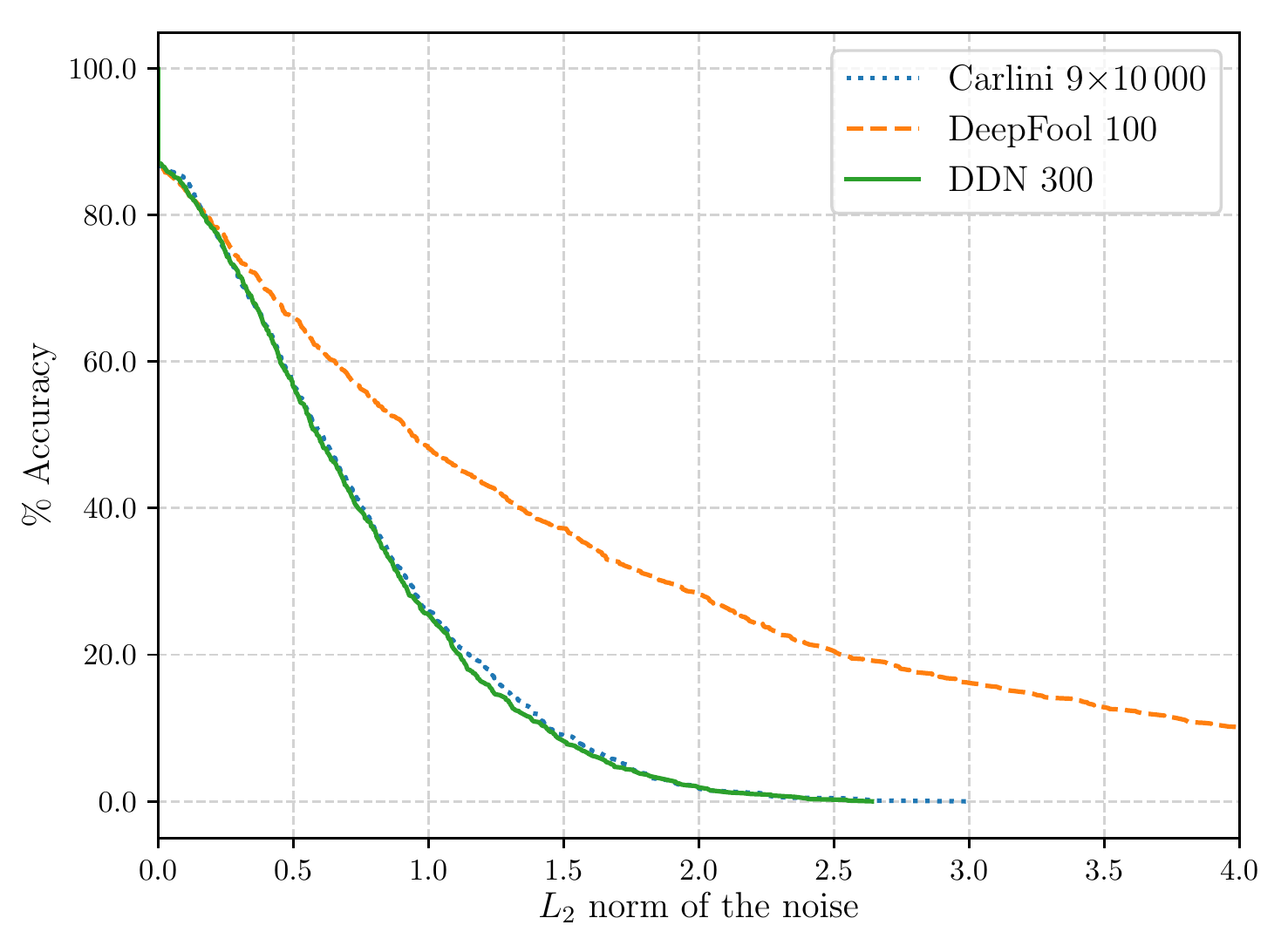}}
    \subfloat[CIFAR-10 / Our Defense.]
    {\includegraphics[width=0.75\columnwidth]{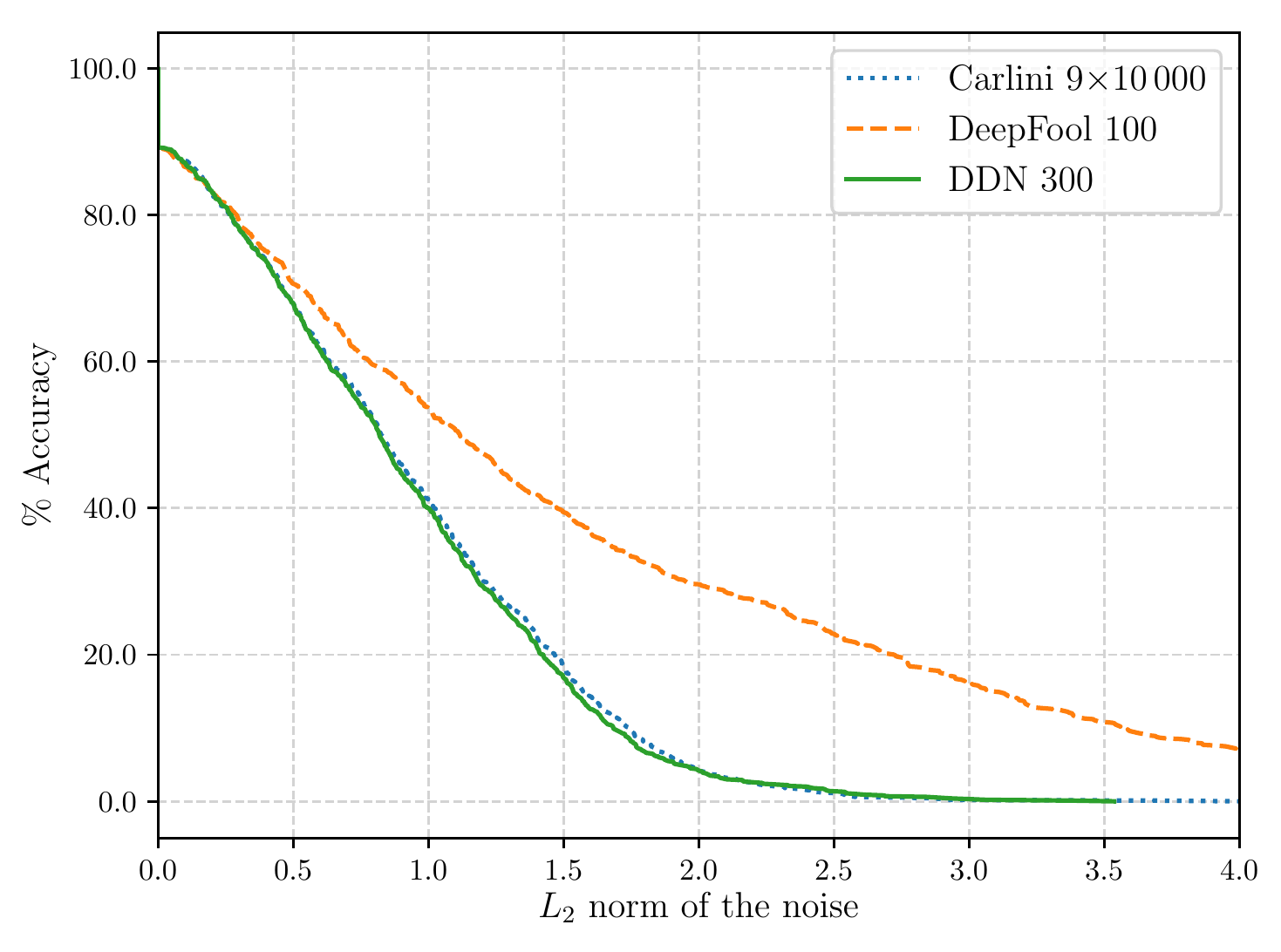}}
    \caption{Attacks performances on different datasets and models.}
    \label{fig:untargeted_attacks}
\end{figure*}

\autoref{fig:untargeted_attacks} reports curves of the perturbation size against accuracy of the models for three attacks: Carlini 9$\times$10\,000, DeepFool 100 and DDN 300.

\end{document}